\documentclass[10pt,twocolumn,letterpaper]{article}
\usepackage[pagenumbers]{cvpr}
\usepackage{graphicx}
\usepackage{amsmath,amssymb,amsthm,mathtools}
\usepackage{booktabs,multirow,array}
\IfFileExists{algorithm.sty}{\usepackage{algorithm}}{%
  \usepackage{float}
  \floatstyle{ruled}
  \newfloat{algorithm}{tbp}{loa}
  \floatname{algorithm}{Algorithm}
  \floatstyle{plain}
}
\IfFileExists{algpseudocode.sty}{\usepackage{algpseudocode}}{%
\newcounter{ridexalgline}
\newlength{\ridexalgindent}
\newenvironment{algorithmic}[1][1]{%
  \setcounter{ridexalgline}{0}\setlength{\ridexalgindent}{0pt}%
  \begin{list}{}{%
    \setlength{\leftmargin}{1.7em}\setlength{\labelwidth}{1.2em}%
    \setlength{\labelsep}{0.5em}\setlength{\itemsep}{0pt}%
    \setlength{\topsep}{0pt}\setlength{\partopsep}{0pt}%
    \setlength{\parsep}{0pt}}%
}{\end{list}}
\newcommand{\Require}{\item[\textbf{Require:}]}
\newcommand{\State}{%
  \stepcounter{ridexalgline}%
  \item[\footnotesize\arabic{ridexalgline}:]\hspace*{\ridexalgindent}\ignorespaces}
\newcommand{\For}[1]{\State\textbf{for} #1 \textbf{do}\addtolength{\ridexalgindent}{1.5em}}
\newcommand{\EndFor}{\addtolength{\ridexalgindent}{-1.5em}\State\textbf{end for}}
\newcommand{\If}[1]{\State\textbf{if} #1 \textbf{then}\addtolength{\ridexalgindent}{1.5em}}
\newcommand{\Else}{\addtolength{\ridexalgindent}{-1.5em}\State\textbf{else}\addtolength{\ridexalgindent}{1.5em}}
\newcommand{\EndIf}{\addtolength{\ridexalgindent}{-1.5em}\State\textbf{end if}}
\newcommand{\Comment}[1]{\hfill\(\triangleright\) #1}

}
\usepackage{tikz}
\usetikzlibrary{arrows.meta,positioning,calc}
\usepackage{microtype}
\usepackage{xspace}
\usepackage{url}
\definecolor{cvprblue}{rgb}{0.21,0.49,0.74}
\usepackage[pagebackref,breaklinks,colorlinks,allcolors=cvprblue]{hyperref}
\newcommand{\method}{RiDEx\xspace}
\newcommand{\E}{\mathbb{E}}
\newcommand{\R}{\mathbb{R}}
\newcommand{\sg}{\operatorname{sg}}
\newcommand{\norm}[1]{\left\lVert #1\right\rVert}

\newcommand{\Ldir}{\mathcal L_{\mathrm{dir}}}
\newcommand{\Lnft}{\mathcal L_{\mathrm{NFT}}}

\DeclareMathOperator{\clip}{clip}

\def\confName{CVPR}
\def\confYear{2027}

\title{RiDEx: Reward-Informed Directional Exploration\\for Diffusion Reinforcement Learning}
\author{%
Song Yan$^{1}$\thanks{Email: \texttt{gary\_144@mail.ustc.edu.cn}. $^{\dagger}$Corresponding authors.} \quad
Chenfeng Wang$^{1,2}$ \quad
Wei Zhai$^{1\dagger}$ \quad
Xingchen Wang$^{1}$ \quad
Tao Zhang$^{3}$ \\
Yawen Shao$^{1,4}$ \quad
Shi Pan$^{5}$ \quad
Min Li$^{5\dagger}$ \quad
Zheng-Jun Zha$^{1}$ \\[4pt]
{\small $^{1}$University of Science and Technology of China \quad $^{2}$Li Auto Inc. \quad $^{3}$Huazhong University of Science and Technology} \\
{\small $^{4}$Tongyi Lab \quad $^{5}$Xi'an High-tech Research Institute} \\[4pt]
{\small Project Page: \url{http://gary.ac.cn/RiDEX/}}
}
\begin{document}
\maketitle
\begin{abstract}
Reward-based diffusion post-training updates model parameters from task rewards. While existing approaches improve sampling and policy optimization, they still rely largely on isotropic random perturbations in a high-dimensional latent space for exploration. Such variation often spends the perturbation budget on changes with little influence on the target reward, making exploration inefficient. We study where diffusion models should explore and how to learn from that exploration. We introduce Reward-Informed Directional Exploration (RiDEx), which concentrates local exploration in reward-relevant subspaces of the denoising velocity field. Conditional residuals or differentiable reward gradients propose the subspace, and equal-and-opposite perturbations at the same generation state measure its terminal reward response. The reward difference supplies signed local supervision in the explored coordinates, while all generated images contribute to global DiffusionNFT learning. RiDEx thereby couples exploration-space selection with feedback about the changes it produces. On Stable Diffusion 3.5 Medium, RiDEx attains the highest peak rewards among the evaluated methods on four alignment and preference tasks. Its gradient-direction instantiation achieves $15\times$ and $29\times$ target-attainment speedups over DiffusionNFT on HPSv2 and PickScore, measured in H200 GPU hours. Controlled experiments confirm that reward-aligned directions elicit stronger responses at equal perturbation magnitude, and that applying their measured credit improves curve-average reward across all four tasks.
\end{abstract}

\section{Introduction}
\begin{figure*}[t]
\centering
\includegraphics[width=\textwidth]{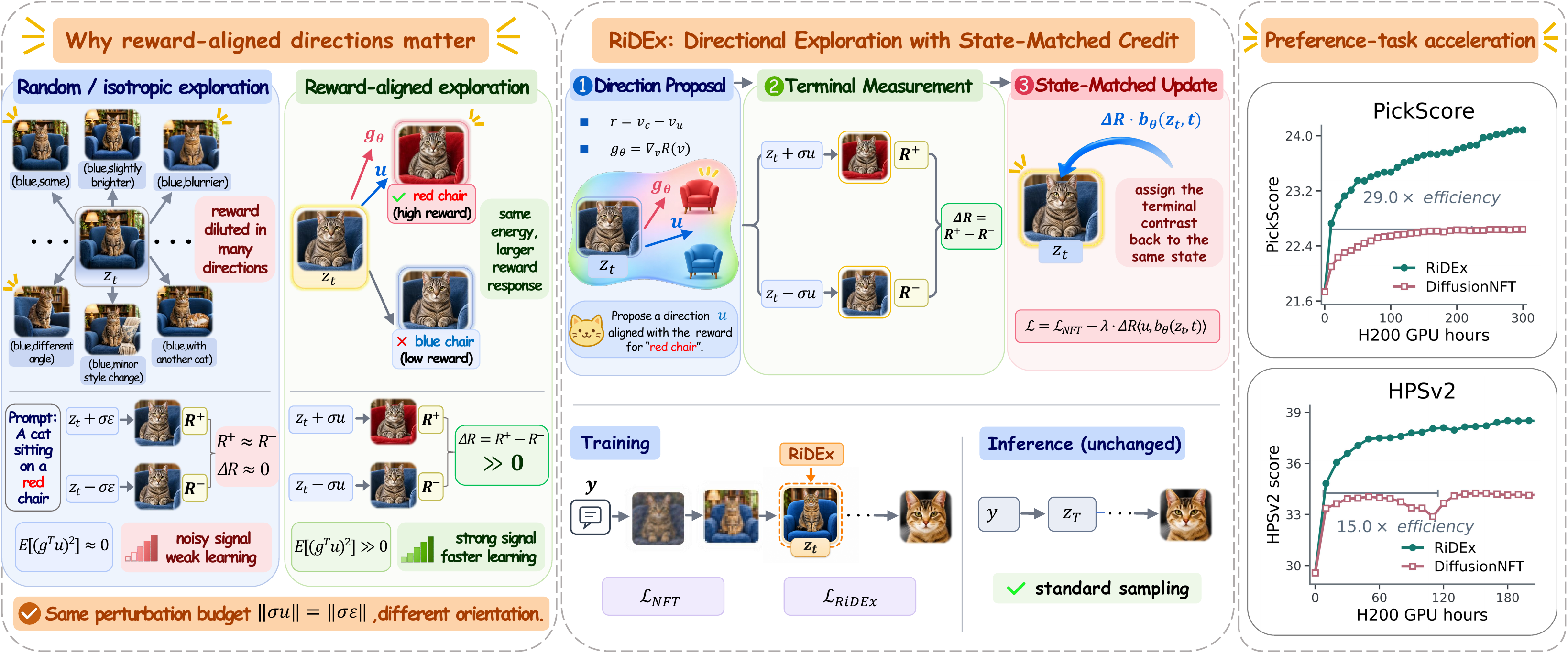}
\caption{\textbf{Reward-informed exploration turns targeted changes into learning signals.} \emph{Left:} At equal perturbation norm, unrelated variations leave the chair blue, whereas a reward-relevant change targets the requested color. \emph{Center:} RiDEx proposes directions, scores paired endpoints, and supervises their originating state. \emph{Right:} Preference target attainment in H200 GPU hours. The schematic depicts RiDEx-Grad; Eqs.~\ref{eq:action},~\ref{eq:credit}, and~\ref{eq:cfg-update} define the interventions, normalized credit, and CFG update.}
\label{fig:method}
\end{figure*}

Reward-based post-training adapts pretrained diffusion models to downstream objectives that visual realism alone cannot guarantee: accurate object relations, legible scene text, and alignment with nuanced human preferences~\cite{lee2023aligning,black2023training,fan2023dpok}. Because every reward query requires unrolling a multi-step denoising trajectory and evaluating the decoded image, post-training efficiency hinges on a fundamental question: how much task-relevant information does each explored trajectory actually reveal?

Recent work has substantially improved key stages of this optimization pipeline. Flow-GRPO extends policy optimization to stochastic denoising transitions~\cite{liu2025flow}; temporal allocation methods concentrate rollout budgets on critical denoising steps~\cite{he2026tempflow,li2025mixgrpo}; DiffusionNFT supervises re-noised endpoints directly through the forward diffusion process~\cite{zheng2026nft}; and ExploreNet adapts coordinate noise scales from observed rewards~\cite{li2026explorenet}. Yet while these advances accelerate trajectory sampling and stabilize policy updates, they still treat exploration largely as undirected noise injection, leaving open a core question: \emph{which semantic variations are actually worth exploring?}

Standard post-training exploration perturbs initial latents or intermediate denoising states with isotropic Gaussian noise. In a high-dimensional latent space, however, an equal-norm random vector is nearly orthogonal to any specific reward-sensitive direction, as its expected squared projection onto a target gradient scales inversely with ambient dimension. A random rollout can therefore alter an image visibly while yielding almost no signal on how to improve its task score. Figure~\ref{fig:method} illustrates this mismatch: for \emph{``a cat sitting on a red chair,''} drawing multiple random rollouts varies lighting, viewpoint, or the cat's pose while leaving the chair stubbornly blue ($\Delta R \approx 0$), whereas perturbing along the color-relevant direction directly tests the requested attribute. The core bottleneck is thus how exploration energy is allocated to target-relevant changes, not merely how many random rollouts are drawn.

Overcoming this bottleneck requires concentrating variation on reward-sensitive changes and learning directly from their measured effects. Crucially, simply restricting exploration to an arbitrary low-dimensional or generic feature subspace is not enough; the chosen subspace must capture strong target-reward response. This motivates our central question: \emph{In which space should diffusion models explore, and how should exploration in that space guide policy learning?}

To answer this, we introduce \method, \emph{Reward-Informed Directional Exploration}, which confines local exploration to a task-informed subspace of the model's denoising velocity field (the vector field transporting latents across steps). Because post-training tasks differ in how reward signals can be accessed, \method instantiates subspace selection according to reward differentiability: for derivative-free text--image alignment objectives, \textbf{RiDEx-CFG} anchors a compact two-dimensional subspace using the model's own conditional velocity residual; when a differentiable reward model is available, \textbf{RiDEx-Grad} backpropagates through the scorer to follow the target-specific reward gradient. Equal-and-opposite velocity perturbations at a shared state then produce two completed images whose terminal reward difference sets both the sign and strength of local supervision in the explored coordinates, tying the learning signal directly to the action that produced it.

These paired endpoints simultaneously enter the global DiffusionNFT objective alongside unperturbed clean trajectories, so each pair supplies both reward-scored training images and a controlled local comparison at a shared state. \method integrates this local directional credit with global policy learning through norm-matched orthogonal projection (RiDEx-CFG) or direct gradient addition (RiDEx-Grad), unifying subspace selection, terminal reward measurement, and policy improvement within a single framework.

On SD3.5 Medium, \method attains the highest peak rewards among all evaluated methods on GenEval, OCR, HPSv2, and PickScore, with RiDEx-Grad reaching DiffusionNFT's HPSv2 and PickScore targets $15\times$ and $29\times$ faster in H200 GPU hours. Controlled frozen-policy interventions confirm that reward response vanishes as perturbations rotate away from target-aligned directions at fixed energy, while applying measured local credit consistently elevates learning curves across all four tasks. These findings establish \textbf{choosing where to explore and how to learn from it} as a principled route to sample-efficient diffusion post-training.

\section{Background}
\subsection{Diffusion and Flow Models}
Diffusion sampling admits deterministic and stochastic formulations~\cite{song2021ddim,song2021sde}. For clean latent $z_0$, noise $\epsilon$, prompt $c$, and $t\in[0,1]$, flow matching and rectified flow~\cite{lipman2022flow,liu2022rectified} use
\begin{equation}
 z_t=(1-t)z_0+t\epsilon,\qquad
 \mathcal L_{\mathrm{FM}}=\E\norm{v_\theta(z_t,t,c)-(\epsilon-z_0)}^2.
 \label{eq:flow}
\end{equation}
Sampling integrates from noise to data before a frozen VAE decodes the endpoint; we denote the current and behavior models by $v_\theta$ and $v_{\bar\theta}$. While classifier guidance uses a classifier gradient~\cite{dhariwal2021guided}, CFG combines conditional and unconditional predictions~\cite{ho2022classifier}. RiDEx uses this conditional residual as a derivative-free source of exploration directions.

\subsection{Forward-Process Reward Updates}
PPO~\cite{schulman2017ppo} and group-relative policy optimization~\cite{shao2024grpo} provide foundations for online reward learning. Whereas Flow-GRPO uses stochastic transition likelihoods~\cite{liu2025flow}, DiffusionNFT learns from re-noised endpoints~\cite{zheng2026nft}. We retain NFT's velocity branches,
\begin{equation}
 v_\theta^+=\beta v_\theta+(1-\beta)v_{\bar\theta},\qquad
 v_\theta^-=(1+\beta)v_{\bar\theta}-\beta v_\theta,
 \label{eq:nft-branches}
\end{equation}
and its reward-weighted reconstruction objective with group-centered rewards, round-wide standard deviation, and a pretrained-velocity penalty (supplement). While NFT supplies global sample-level supervision from endpoint rewards, RiDEx adds a controlled local comparison: selecting a velocity subspace, measuring the reward change within it, and supervising the corresponding velocity coordinates.

\section{Reward-Informed Directional Exploration}
\method couples targeted exploration-space selection with state-matched supervision learned from the resulting terminal reward response. Its shared interface comprises a state-dependent orthonormal basis $U_h$, a coordinate-credit estimator $\mathcal E_h$, and a gradient-combination rule $\mathcal A$. \textbf{RiDEx-CFG} and \textbf{RiDEx-Grad} instantiate this framework using derivative-free conditional residuals and differentiable reward feedback, respectively.

\subsection{Selecting the Exploration Space}
\label{sec:action}
A generation center $h$ encapsulates the prompt $c$, initial noise, latent $z_\tau$, time $\tau$, and solver history. Let $v_h=v_{\bar\theta}(z_\tau,\tau,c)$ denote the behavior velocity at center $h$ with scale $q_h=\norm{v_h}_2>0$. An orthonormal basis $U_h\in\R^{D\times k}$ spans a local subspace $\mathcal S_{h,R}=\operatorname{range}(U_h)$ in velocity-output space ($k\ll D$). For a unit coefficient vector $s\in\R^k$ and relative radius $r>0$, we branch one velocity prediction into symmetric probes:
\begin{equation}
 v_h^{\mathrm{inj}}(a)=v_h+q_h a,\qquad
 a_\pm=\pm rU_hs,\quad U_h^\top U_h=I_k.
 \label{eq:action}
\end{equation}
Scaling by $q_h$ fixes relative action magnitude independently of dimension $k$, while $U_h$ governs \emph{where} that perturbation budget is spent. Let $F_h(a)$ be the completed image under the frozen continuation and $f_h(a)=R(F_h(a),c)$ its reward. Because both branches share the exact prefix, center $h$, and solver schedule---reconstructed at the first step or replayed from a saved intermediate state---any terminal reward difference is causally tied to the local action $\pm rU_hs$.

\paragraph{Selecting response, not dimension alone.}
\label{sec:geometry}
For differentiable $f_h$ with action gradient $g_h=\nabla_a f_h(0)$, isotropic unit directions in the ambient space $\R^D$ and subspace $\mathcal S_{h,R}$ satisfy
\begin{equation}
 \E_u(g_h^\top u)^2=\frac{\norm{g_h}^2}{D},\qquad
 \E_s(g_h^\top U_hs)^2=\frac{\norm{U_h^\top g_h}^2}{k}.
 \label{eq:random-response}
\end{equation}
Their ratio equals $(D/k)\rho_h$, where $\rho_h=\norm{U_h^\top g_h}^2/\norm{g_h}^2$ is the fraction of gradient energy captured by $U_h$ ($g_h\ne0$). Restricting exploration to $U_h$ thus amplifies expected squared response whenever $\rho_h>k/D$, capturing more reward sensitivity per coordinate~\cite{constantine2014active}. Since an independent isotropic subspace has $\E[\rho_h]=k/D$, dimensionality reduction alone yields no expected gain. For smooth rewards, the paired contrast $2r g_h^\top U_hs+O(r^3)$ links subspace alignment directly to terminal measurements while remaining defined for discrete rewards.

\begin{figure*}[t]
\centering
\includegraphics[width=\textwidth]{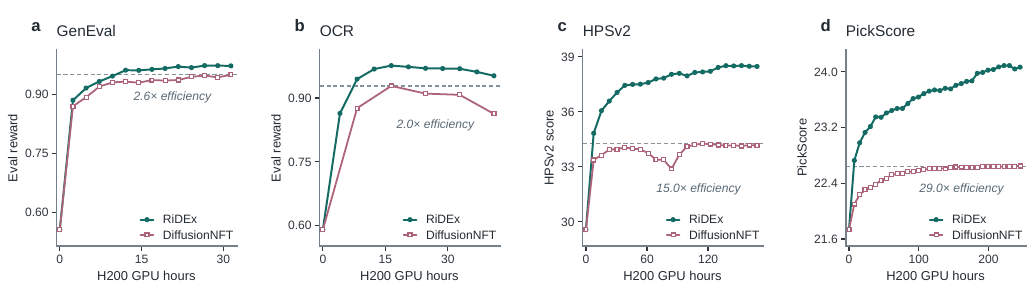}
\caption{\textbf{Reward-efficient learning in H200 GPU hours.} Dashed lines mark the shared target in each panel, with attainment measured at the first checkpoint reaching that score. Table~\ref{tab:attainment} gives the corresponding GPU times. RiDEx denotes RiDEx-CFG on GenEval and OCR and RiDEx-Grad on HPSv2 and PickScore.}
\label{fig:learning}
\end{figure*}

\subsection{Proposing Spaces and Measuring Their Response}
\label{sec:proposals}
This captured-response criterion guides proposal construction across reward regimes. For rule-based or detector-driven text--image alignment rewards (e.g., GenEval and OCR), reward derivatives are unavailable, but the diffusion model itself encodes prompt-responsive variations via its conditional residual. \textbf{RiDEx-CFG} leverages this prior by anchoring a two-dimensional conditional basis at the first step around
\begin{equation}
 p_h^{\mathrm{CFG}}=v_h-v_{\bar\theta}(z_\tau,\tau,\varnothing).
 \label{eq:cfg}
\end{equation}
First-step probes $U_h^{\mathrm{CFG}}s$ thus target prompt-relevant semantics without reward derivatives. Conversely, when a differentiable scoring path $\widetilde R$ is available, \textbf{RiDEx-Grad} extracts a target-specific ascent line at a saved intermediate center:
\begin{align}
 p_h^{\mathrm{grad}}&=\left.\nabla_a\widetilde R(F_h(a),c)\right|_{a=0},\nonumber\\
 U_h&=[u_h],\qquad u_h=p_h^{\mathrm{grad}}/\norm{p_h^{\mathrm{grad}}}_2.
 \label{eq:gradient-proposal}
\end{align}
Differentiation traverses the frozen remaining sampler, VAE, and scorer (zero-norm proposals receive zero action). Unlike generic semantic readouts~\cite{kwon2022semantic,park2023riemannian} that capture general visual variance, Eqs.~\ref{eq:cfg}--\ref{eq:gradient-proposal} align $U_h$ with the target objective.

While $U_h$ specifies which directions to test, the canonical evaluator $R$ measures their realized effect over a finite step. For coefficient $s_{hj}$ and radius $r_{hj}$, let $R_{hj}^{\pm}=f_h(\pm r_{hj}U_hs_{hj})$. The centered finite-difference slope and coordinate credit are
\begin{equation}
 d_{hj}=\frac{R_{hj}^+-R_{hj}^-}{2r_{hj}},
 \quad \widehat c_h=\mathcal E_h(\{s_{hj},d_{hj}\}_j).
 \label{eq:credit}
\end{equation}
RiDEx-CFG aggregates paired finite differences across a balanced 2D design in the shared basis, whereas RiDEx-Grad ($k=1, s=1$) sets $\widehat c_h=d_h$ as signed two-point feedback along the gradient line~\cite{spall1992spsa,nesterov2017random,salimans2017es,maheswaranathan2019guided}. Setting $o_h=(R_+-R_-)/2$ and $e_h=(R_++R_-)/2-R_0$ gives $R_\pm-R_0=\pm o_h+e_h$, separating directional contrast $o_h$ from net gain over clean generation $R_0$.

\subsection{Learning from the Explored Changes}
\label{sec:local-supervision}
The measured credit $\widehat c_h$ directly supervises the velocity prediction at the originating state $h$. Defining the normalized policy displacement $b_\theta(h)$, the local directional loss is
\begin{align}
 b_\theta(h)&=\frac{v_\theta(z_\tau,\tau,c)-v_h}{q_h},\nonumber\\
 \Ldir&=-\frac1M\sum_h\sg[\widehat c_h]^\top
                  \sg[U_h]^\top b_\theta(h).
 \label{eq:local-loss}
\end{align}
Detaching collection quantities ($\sg$) pulls the coordinate credit back through the originating state's velocity Jacobian:
\begin{equation}
 -\nabla_\theta\Ldir=\frac1M\sum_h\frac1{q_h}
  (\nabla_\theta v_\theta(h))^\top U_h\widehat c_h.
 \label{eq:pullback}
\end{equation}
Exploration, terminal measurement, and local supervision thus refer to the exact same state $h$ and coordinate frame $U_h$.

\textbf{Endpoint scores determine how strongly to learn.} RiDEx-Grad uses its gradient to select the line and scored endpoints to weight it: a tie yields zero credit, and a better negative branch favors $-u_h$. By averaging directional response across $[-r,r]$ under the canonical evaluator, paired slopes provide curvature-aware state weights whose $r\to0$ limit recovers the center slope (supplement, Sec.~A).

\begin{figure*}[t]
\centering
\includegraphics[width=\textwidth]{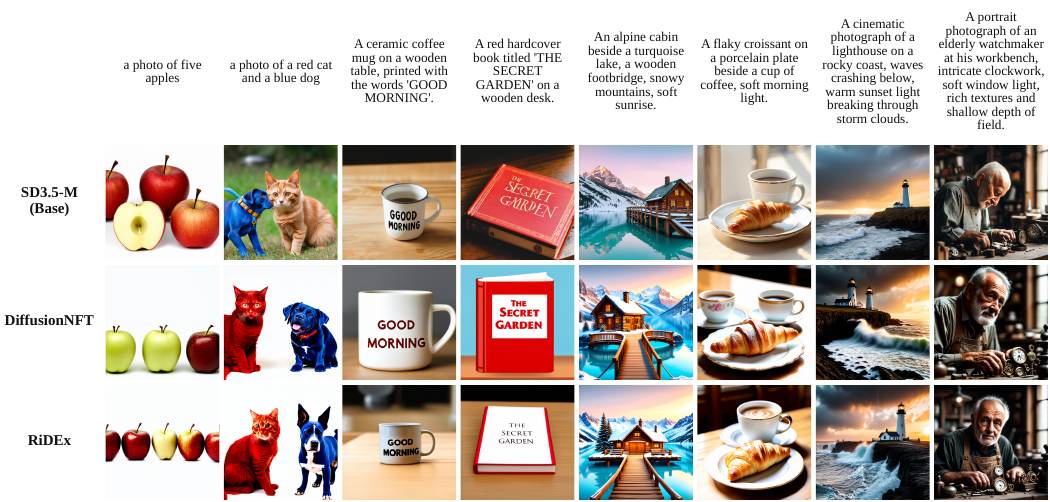}
\caption{\textbf{Qualitative comparison across alignment and preference benchmarks.} Columns span GenEval (counting and color binding), OCR (text legibility in natural scenes), HPSv2, and PickScore prompts, comparing pretrained SD3.5-M, DiffusionNFT, and RiDEx.}
\label{fig:qualitative}
\end{figure*}

\subsection{Combining Local Credit with Global Learning}
\label{sec:update-rules}
While $\Ldir$ supervises specific local interventions at center $h$, the global DiffusionNFT objective $\Lnft$ trains the velocity network across all re-noised timesteps from complete endpoint images. Let $G_N=\nabla_\theta\Lnft$ and $G_D=\nabla_\theta\Ldir$ denote their parameter gradients at current parameters $\theta$. Each instantiation merges $G_N$ and $G_D$ according to the properties of its credit estimate before joint clipping and optimization.

\paragraph{RiDEx-Grad: direct combination.}
Because RiDEx-Grad measures signed credit along a reward-gradient line in canonical reward units, its local gradient $G_D$ carries both informative orientation and calibrated state-wise magnitude:
\begin{equation}
 G_{\mathrm{Grad}}=G_N+\lambda G_D
 =\nabla_\theta(\Lnft+\lambda\Ldir),
 \label{eq:joint}
\end{equation}
allowing local feedback to refine both the direction and scale of the joint parameter update.

\paragraph{RiDEx-CFG: projected combination.}
In derivative-free alignment tasks, two-dimensional finite differences supply directional guidance across discrete reward levels. RiDEx-CFG therefore injects the complementary component of $G_D$ orthogonal to $G_N$ while preserving the global NFT gradient norm. For nonzero $G_N$ and $G_D^\perp$,
\begin{align}
 G_D^\perp&=G_D-\frac{\langle G_D,G_N\rangle}{\norm{G_N}^2}G_N,\nonumber\\
 H&=G_N+\eta\frac{\norm{G_N}}{\norm{G_D^\perp}}G_D^\perp,\qquad
 G_{\mathrm{CFG}}=\frac{\norm{G_N}}{\norm{H}}H.
 \label{eq:cfg-update}
\end{align}
The effective trust coefficient $\eta$ incorporates confidence weighting (reducing cleanly to $G_N$ at zero local signal). Orthogonality lets $|\eta|$ directly control the added direction's relative length, while norm matching keeps $\norm{G_{\mathrm{CFG}}}=\norm{G_N}$. Crucially, this projection acts on \emph{parameter gradients}, whereas $U_h$ selects variations in \emph{velocity-output space}.

\subsection{Global Coverage with Targeted Probes}
Each training round combines independent clean trajectories with symmetric local probes: all completed endpoints train $\Lnft$ for broad trajectory coverage, while each branched pair additionally supplies state-matched coordinate supervision via $\Ldir$. Both signals jointly refine a single velocity field $v_\theta$, deployed at test time via standard sampling without paired search (supplement).

\section{Experiments}
We evaluate training efficiency and use controlled interventions to link subspace selection, terminal response, and policy learning.
\subsection{Tasks and Comparisons}
\label{sec:setup}
We evaluate SD3.5M~\cite{stability2024sd35} on four rewards. GenEval~\cite{ghosh2023geneval} scores compositional agreement and OCR measures text legibility; both are non-differentiable alignment tasks instantiated with \textbf{RiDEx-CFG} (2D conditional subspace and projected norm-matched update). HPSv2~\cite{wu2023hpsv2} and PickScore~\cite{kirstain2023pick} are differentiable preference scorers instantiated with \textbf{RiDEx-Grad} (1D reward-gradient proposal and direct gradient addition; Sec.~\ref{sec:update-rules}). We measure H200 GPU hours to a shared reward target, followed by frozen-policy probes and credit ablations (protocols in the supplement).

\begin{table}[t]
\centering\small\setlength{\tabcolsep}{3.5pt}
\begin{tabular}{@{}lrrrr@{}}
\toprule
Method & GenEval & OCR & HPSv2 & PickScore\\
\midrule
Flow-GRPO~\cite{liu2025flow} & 0.9500 & 0.9200 & 31.6000 & 23.5000\\
ReFL~\cite{xu2023imagereward} & --- & --- & 35.8073 & 23.9211\\
ExploreNet-GRPO~\cite{li2026explorenet} & --- & --- & 30.1798 & 22.1234\\
DiffusionNFT~\cite{zheng2026nft} & 0.9501 & 0.9285 & 34.2538 & 22.6407\\
RiDEx & \textbf{0.9733} & \textbf{0.9683} & \textbf{38.5154} & \textbf{24.0847}\\
\bottomrule
\end{tabular}
\caption{\textbf{Reward scores on SD3.5 Medium.} Bold marks the highest peak reward for each task. HPSv2 is scaled by 100; PickScore is reported in logit units.}
\label{tab:learning}
\end{table}

\begin{table}[t]
\centering\small\setlength{\tabcolsep}{3.8pt}
\begin{tabular}{lrrrr}
\toprule
Task & Target & NFT (h) & RiDEx (h) & Speedup\\
\midrule
GenEval & 0.9501 & 31.39 & 12.07 & $2.6\times$\\
OCR & 0.9285 & 16.44 & 8.22 & $2.0\times$\\
HPSv2 & 34.2538 & 114.63 & 7.64 & $15.0\times$\\
PickScore & 22.6407 & 222.52 & 7.67 & $29.0\times$\\
\bottomrule
\end{tabular}
\caption{\textbf{H200 GPU hours to matched reward targets.} Approximate times are read from the target-attaining Figure~\ref{fig:learning} checkpoints; speedups use unrounded coordinates. The same target is applied to both methods.}
\label{tab:attainment}
\end{table}

\subsection{Reward-Efficient Policy Learning}
\label{sec:learning}
\textbf{RiDEx achieves the highest peak reward on every evaluated task.} Table~\ref{tab:learning} and Figure~\ref{fig:learning} compare RiDEx with Flow-GRPO, ReFL, ExploreNet-GRPO, and DiffusionNFT. RiDEx surpasses all baselines across alignment and preference tasks, improving over DiffusionNFT by $+0.0232$ on GenEval, $+0.0398$ on OCR, $+4.2616$ on HPSv2, and $+1.4440$ on PickScore, while also exceeding ReFL by $+2.7081$ HPSv2 and $+0.1636$ PickScore points. Moreover, RiDEx leads DiffusionNFT at all shared checkpoints on GenEval (13/13), HPSv2 (21/21), and PickScore (29/29).

\paragraph{GPU time to a common target.}
For method $m$, baseline peak $\tau$, and checkpoint $j$ at $H_{m,j}$ H200 GPU hours, we report
\begin{equation}
 T_m(\tau)=\min_{j:\,R_{m,j}\ge\tau}H_{m,j},\qquad
 S(\tau)=\frac{T_{\mathrm{NFT}}(\tau)}{T_{\mathrm{RiDEx}}(\tau)}.
 \label{eq:time-to-target}
\end{equation}
RiDEx-Grad reaches the HPSv2 and PickScore targets in $7.64$ and $7.67$ GPU hours versus $114.63$ and $222.52$ for DiffusionNFT ($15.0\times$ and $29.0\times$ speedups), while RiDEx-CFG accelerates GenEval and OCR attainment by $2.6\times$ and $2.0\times$ (Table~\ref{tab:attainment}).

\begin{figure*}[t]
\centering
\captionsetup{skip=6pt}
\includegraphics[width=\textwidth]{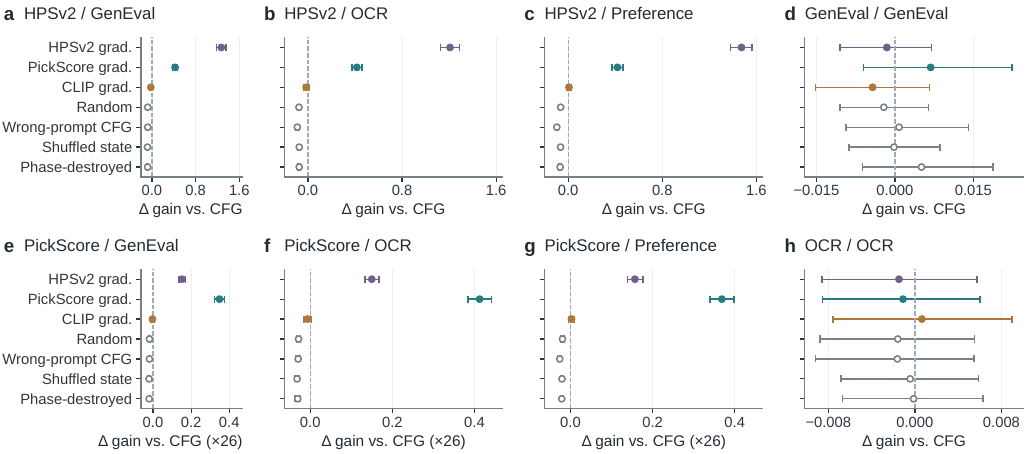}
\caption{\textbf{The exploration space must match the target reward.} Seven directions are compared with same-state CFG at equal radius and query count. Zero marks the CFG response. Points and bars give the prompt-cluster mean and its 95\% interval.}
\label{fig:directions}
\end{figure*}

\begin{figure*}[t]
\centering
\captionsetup{skip=6pt}
\includegraphics[width=\textwidth]{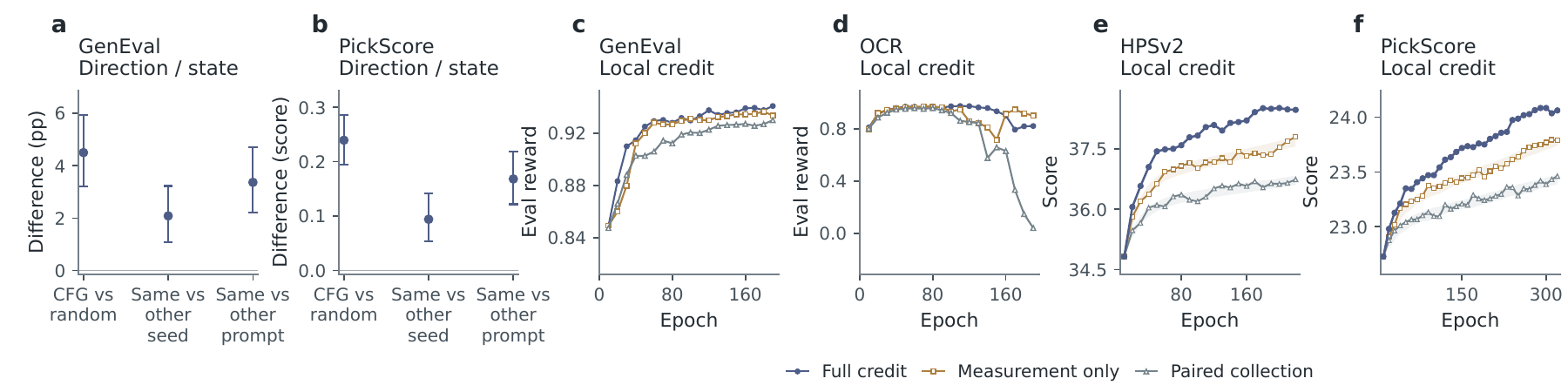}
\caption{\textbf{Local learning depends on direction, state, and credit application.} (a,b) Held-out branch advantage when changing the space or the source of its coefficients (95\% bootstrap intervals). (c--f) Learning curves comparing full credit (circles), measurement only (squares), and paired collection (triangles) across training epochs, summarized in Table~\ref{tab:credit-control}.}
\label{fig:cfgcredit}
\end{figure*}

\paragraph{Qualitative comparisons.}
Figure~\ref{fig:qualitative} shows how targeted exploration improves visual generation across all four tasks. On GenEval prompts, Base and DiffusionNFT miscount objects (\emph{``five apples''}) or misbind colors (\emph{``a red cat and a blue dog''}), whereas RiDEx satisfies exact counts and color bindings with clean object separation. On OCR prompts, Base misspells text (\emph{``GGOOD''} or garbled book titles) and DiffusionNFT resorts to cropped or flat synthetic backgrounds, while RiDEx renders crisp, accurate lettering within realistic tabletop scenes. On HPSv2 and PickScore prompts, RiDEx eliminates structural artifacts such as DiffusionNFT's extra coffee cup and duplicate lighthouse tower, yielding coherent scene geometry, natural sunrise and coastal lighting, and rich portrait textures.

\subsection{Which Spaces Are Reward-Relevant?}
\label{sec:direction-results}
To isolate subspace orientation before parameter updates, we probe 1D lines $\operatorname{span}\{u_h\}$ at equal state and radius $r$, comparing the CFG residual, HPSv2, PickScore, and CLIP~\cite{radford2021clip} gradients, plus random, wrong-prompt CFG, shuffled, and phase-randomized controls against same-state CFG:
\begin{equation}
 \Delta_h(u)=[f_h(ru)-f_h(0)]-
 [f_h(ru_{\mathrm{CFG}})-f_h(0)].
 \label{eq:contrast-to-cfg}
\end{equation}
\textbf{Exploration subspaces must match the target reward.} In Figure~\ref{fig:directions}, reward-matched gradients strongly outperform CFG across all prompt domains ($+1.47$ HPSv2 and $+0.37$ PickScore points on preference prompts), whereas random, shuffled, and phase-destroyed probes yield near-zero response. Within every domain, HPSv2 favors its own gradient over PickScore's while PickScore shows the reverse ordering, confirming that optimal exploration directions are objective-specific. On GenEval and OCR, CLIP-gradient responses closely match CFG ($\Delta_h\approx0$), supporting RiDEx-CFG's derivative-free conditional anchor.

\begin{figure*}[t]
\centering
\captionsetup{skip=6pt}
\includegraphics[width=\textwidth]{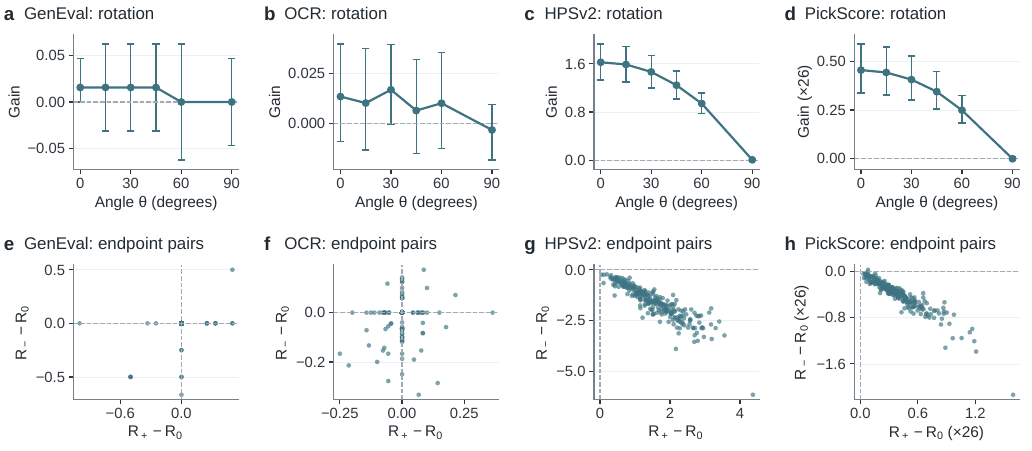}
\caption{\textbf{Reward relevance is visible at the endpoint.} (a--d) Rotation toward an orthogonal direction at fixed radius eliminates reward gain. (e--h) Rewards of 1,024 branch pairs relative to their clean endpoint $R_0$. Bars show 95\% prompt-cluster intervals.}
\label{fig:finite}
\end{figure*}

\begin{table}[t]
\centering\small\setlength{\tabcolsep}{3pt}
\begin{tabular}{@{}lrrrr@{}}
\toprule
Arm & GenEval & OCR & HPSv2 & PickScore\\
\midrule
Paired collection & 0.9123 & 0.7640 & 36.3102 & 23.2071\\
Measurement only & 0.9206 & 0.9100 & 37.0169 & 23.4473\\
Full credit & \textbf{0.9260} & \textbf{0.9305} & \textbf{37.7607} & \textbf{23.6770}\\
\bottomrule
\end{tabular}
\caption{\textbf{Mean AUC for the credit ablation.} For each task, the trapezoidal area under the reward curve in Fig.~\ref{fig:cfgcredit} is divided by the length of the evaluation window shared by the three arms.}
\label{tab:credit-control}
\end{table}

\begin{table}[t]
\centering\small\setlength{\tabcolsep}{3pt}
\begin{tabular}{llrrr}
\toprule
Target & Direction & Odd $o_h$ & Even $e_h$ & Win+ (\%)\\
\midrule
HPSv2 & Own & \textbf{1.5624} & -0.0240 & \textbf{100.0}\\
 & CFG & 0.0674 & -0.0017 & 87.5\\
 & Random & 0.0023 & -0.0005 & 55.1\\
 & Shuffled & 0.0017 & -0.0008 & 52.0\\
\midrule
PickScore & Own & \textbf{0.3916} & -0.0004 & \textbf{100.0}\\
 & CFG & 0.0215 & 0.0011 & 81.2\\
 & Random & 0.0010 & 0.0020 & 57.4\\
 & Shuffled & 0.0005 & 0.0013 & 53.1\\
\bottomrule
\end{tabular}
\caption{\textbf{Reward-matched positive probes improve every tested preference endpoint.} Means over 256 states per target. Win+ is $\Pr(R_+>R_0)$ under the target evaluator. The odd term measures directional contrast, the even term measures the shared clean-relative response, and their sum gives the positive-branch gain.}
\label{tab:paired}
\end{table}

\begin{figure}[t]
\centering
\captionsetup{skip=4pt}
\includegraphics[width=\columnwidth]{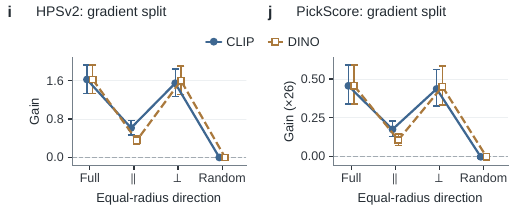}
\caption{\textbf{Reward relevance vs.\ readout sensitivity.} Continuing Fig.~\ref{fig:finite}(a--h), panels (i,j) compare full ($g$), readout-parallel ($\parallel$), complement ($\perp$), and random directions at equal radius for CLIP (circles) and DINOv2 (squares).}
\label{fig:finite-ij}
\end{figure}

\subsection{Direction, State, and Credit Application}
\label{sec:cfg-evidence}
\textbf{The conditional anchor and state specificity are both critical.} When a linear predictor fitted from 2D probes selects the winning sign of held-out test pairs (Fig.~\ref{fig:cfgcredit}(a,b)), the CFG-anchored subspace beats a random 2D subspace by $+4.50$ percentage points (pp) on GenEval. Keeping fitted coefficients at their originating state also outperforms transferring them from another latent ($+2.08$ pp) or another prompt ($+3.36$ pp), with parallel gains on PickScore---confirming that directional credit is state-specific rather than prompt-wide.

\textbf{Applying local credit improves policy learning.} Comparing \emph{paired collection}, \emph{measurement only}, and \emph{full credit} during training (Fig.~\ref{fig:cfgcredit}(c--f), Table~\ref{tab:credit-control}), full credit achieves the highest curve-average reward on all four tasks. Since measurement-only and full credit score the exact same paired endpoints for NFT, these gains isolate the benefit of pulling measured directional contrasts back to the originating state (Sec.~\ref{sec:update-rules}).

\begin{figure}[t]
\centering
\captionsetup{skip=4pt}
\includegraphics[width=\columnwidth]{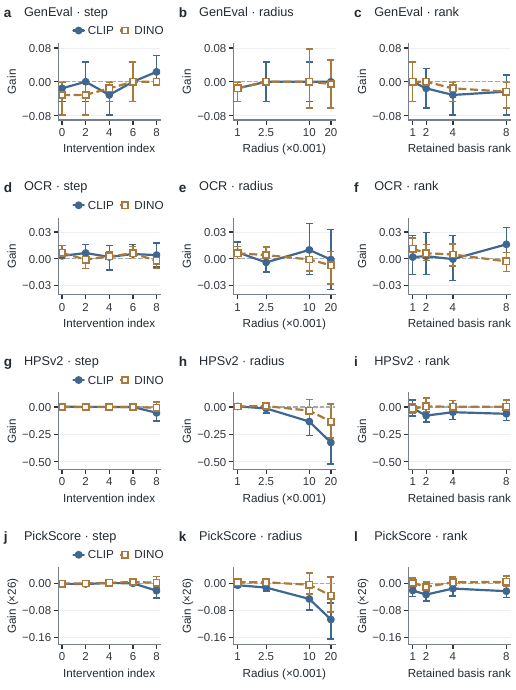}
\caption{\textbf{Energy and rank do not replace target alignment.} Task reward gains across intervention steps (a,d,g,j), radii (b,e,h,k), and retained basis ranks (c,f,i,l) for CLIP (circles) and DINOv2 (squares), with 95\% prompt-cluster intervals.}
\label{fig:sensitivity}
\end{figure}

\subsection{Orientation at Fixed Energy and Dimension}
\label{sec:finite}
To separate orientation from energy, we rotate each task proposal $u_h$ toward an orthogonal unit vector $\xi_h\perp u_h$ at fixed radius:
\begin{equation}
 u_h(\alpha)=\cos\alpha\,u_h+\sin\alpha\,\xi_h,
 \quad\xi_h\perp u_h,\quad\norm{\xi_h}=1.
 \label{eq:angle}
\end{equation}
As $\alpha$ increases from $0^\circ$ to $90^\circ$, reward gain decays monotonically to zero---dropping from $+1.625$ to $+0.005$ on HPSv2 and from $+0.456$ to $-0.002$ on PickScore (Fig.~\ref{fig:finite}(a--d)). \textbf{Perturbation energy without target alignment yields no signal.}

\paragraph{Positive probes improve on clean generation.}
Across all 256 states per preference target, the positive own-gradient branch beats clean generation $R_0$ 100\% of the time (Fig.~\ref{fig:finite}(e--h)). Table~\ref{tab:paired} confirms that directional contrast (odd term $o_h$) accounts for nearly all positive-branch gain, whereas random and shuffled probes yield negligible contrast and near-chance win rates.

\paragraph{Reward relevance differs from readout sensitivity.}
Projecting $g$ onto a finite orthonormal readout basis $U$ (CLIP or DINOv2~\cite{oquab2023dinov2}) and normalizing $g$, $UU^\top g$ ($\parallel$), and $(I-UU^\top)g$ ($\perp$) to equal radius (Fig.~\ref{fig:finite-ij}(i,j)), readout-parallel directions strongly shift visual embeddings yet recover under $40\%$ of the full reward gain, whereas the orthogonal complement retains over $95\%$ ($+1.55/+1.60$ HPSv2 and $+0.44/+0.45$ PickScore). Re-estimating $U$ at each new state also raises held-out response cosine from $0.05\text{--}0.12$ to $0.64\text{--}0.82$ (supplement). Sweeping intervention steps, radii, and retained ranks in Figure~\ref{fig:sensitivity} further confirms that neither larger perturbation radii nor extra generic readout modes can substitute for target alignment (Eq.~\ref{eq:random-response}).

\subsection{From Terminal Response to Policy Supervision}
\label{sec:credit-results}
Finally, we test why RiDEx weights updates by finite-action credit rather than center gradient norms. Ordering $\binom{32}{2}=496$ frozen state pairs by their true reward-gap magnitude at test radius $r'=0.0025$, paired credit measured at $r=0.005$ correctly ranks $481/496$ HPSv2 and $478/496$ PickScore pairs, versus only $136/496$ and $175/496$ for center autodiff norms, with this ranking advantage persisting across all tested radii (supplement, Sec.~C.2). Finite-action measurements thus calibrate relative state importance in a way a global learning-rate scalar cannot replicate. During training, the joint update yields positive credit-aligned displacement in all 226 HPSv2 and 323 PickScore rounds ($>92.7\%$ center-wise sign agreement; supplement, Sec.~D.2), linking informative terminal comparisons directly to realized policy improvement.

\section{Related Work}
\paragraph{Efficient reward optimization.}
DPO~\cite{rafailov2023dpo} and Diffusion-DPO~\cite{wallace2024diffusion} learn from static preference pairs, while DiffusionNFT~\cite{zheng2026nft}, Reinforce Adjoint Matching~\cite{bergmeister2026ram}, and Reward Score Matching~\cite{lee2026rsm} develop online regression or matching objectives. RiDEx retains NFT while reshaping the exploration space that supplies reward contrasts and local supervision.

\paragraph{Exploration distributions and spaces.}
DiffExp adjusts CFG and prompt weights~\cite{chae2025diffexp}, ExploreNet learns diagonal noise scales~\cite{li2026explorenet}, Neighbor GRPO perturbs initial noise~\cite{he2025neighbor}, and RPO explores attention routing~\cite{shen2026rpo}. Principal Steering Subspaces uses response-derived initial-noise spaces for frozen robot policies~\cite{ni2026pss}. RiDEx instead selects target-dependent directions in velocity-output space before querying endpoints.

\paragraph{Semantic guidance and reward relevance.}
CFG++ constrains guidance geometrically~\cite{chung2025cfgpp}, and GuidanceNFT addresses guided/reference velocity mismatch~\cite{liu2026guidancenft}. RiDEx uses the CFG residual as a conditional prior. Across semantic similarity~\cite{hessel2021clipscore}, preference~\cite{kirstain2023pick,wu2023hpsv2,xu2023imagereward}, and compositional faithfulness~\cite{ghosh2023geneval,huang2023t2icompbench,hu2023tifa,cho2023dsg}, our analysis separates reward relevance from readout sensitivity.

\section{Conclusion}
We studied where diffusion models should explore and how that exploration should guide reward-based post-training. RiDEx links both decisions in a shared subspace: task structure proposes candidate changes, paired terminal scores measure their effects, and state-matched supervision turns those measurements into policy updates alongside NFT. Across four alignment and preference tasks, RiDEx achieves the highest peak rewards among evaluated methods, and applying local credit improves curve-average reward. Equal-energy probes further isolate the response of target-aligned directions, while RiDEx-Grad reaches DiffusionNFT's HPSv2 and PickScore targets $15\times$ and $29\times$ faster in H200 GPU time. Allocating exploration to changes that matter to the objective---and preserving their link to learning---thus makes post-training feedback far more effective.

\section*{LLM Usage Statement}
We used GLM for language refinement and code assistance. All generated content and code were carefully reviewed and validated by the authors.
{\small
\bibliographystyle{ieeenat_fullname}
\bibliography{main}
}

\clearpage
\setcounter{page}{1}
\setcounter{figure}{0}
\setcounter{table}{0}
\setcounter{equation}{0}
\renewcommand{\thepage}{S\arabic{page}}
\renewcommand{\thefigure}{S\arabic{figure}}
\renewcommand{\thetable}{S\arabic{table}}
\renewcommand{\theequation}{S\arabic{equation}}
\renewcommand{\theHfigure}{S\arabic{figure}}
\renewcommand{\theHtable}{S\arabic{table}}
\renewcommand{\theHequation}{S\arabic{equation}}
\maketitlesupplementary
\appendix
\section{Mathematical Foundations}
\label{supp:math}
This supplement develops the geometry of reward-informed exploration, specifies the collection and update rules, and links the local mechanism to policy learning. Throughout our empirical analyses, prompt-cluster intervals quantify variation across evaluation prompts.

\subsection{Directional response and captured energy}
Fix a generation center $h$ and let $g=\nabla_a f_h(0)$ denote the local action gradient. For $u$ uniform on the unit sphere in $\R^D$, rotational symmetry gives $\E[uu^\top]=cI_D$. Taking traces yields $1=cD$, and therefore
\begin{equation}
 \E[(g^\top u)^2]=g^\top\E[uu^\top]g=\frac{\norm{g}^2}{D}.
\end{equation}
Similarly, for an orthonormal basis $U^\top U=I_k$ and $\xi$ uniform on the unit sphere in $\R^k$, the same symmetry argument gives
\begin{equation}
 \E[(g^\top U\xi)^2]=\frac{\norm{U^\top g}^2}{k}.
\end{equation}
For $g\ne0$, the response ratio is $(D/k)\rho_U$, with $\rho_U=\norm{U^\top g}^2/\norm{g}^2$. A subspace thus increases expected squared response precisely when $\rho_U>k/D$. For an independently sampled isotropic $k$-dimensional subspace, $\E_U[\rho_U]=k/D$, meaning that dimension reduction alone gives no expected advantage under this criterion; selecting $U$ using reward information is the substantive design choice. These equal-energy identities formalize the distinction between dimensionality and reward relevance. While the identities characterize local response per unit action energy, direction construction, endpoint evaluation, and parameter optimization provide the implementation-level work measures.

\paragraph{Ambient, exploration, and parameter spaces.}
The ambient action space is $\R^D$, with one coordinate per velocity-output entry. RiDEx probes $\mathcal S_{h,R}=\operatorname{range}(U_h)$ with $U_h^\top U_h=I_k$. Its coordinate dimension is two for RiDEx-CFG and one for RiDEx-Grad, varying with both $h$ and $R$. The local objective supervises $U_h^\top b_\theta(h)$, and each center's velocity Jacobian maps its coordinate credit into parameter space. This construction concentrates exploration and local supervision in a shared coordinate system while the model retains its full generative action space.

\subsection{Finite-radius credit and the smooth limit}
For a unit direction $u$, define $\varphi(s)=f_h(su)$ and its centered finite difference
\begin{equation}
 d_r=\frac{\varphi(r)-\varphi(-r)}{2r}.
\end{equation}
If $\varphi$ has a third derivative bounded in magnitude by $M_3$ on $[-r,r]$, Taylor's theorem gives
\begin{equation}
 |d_r-g^\top u|\le M_3r^2/6.
\end{equation}
For a smooth canonical reward, $d_r$ averages $u^\top\nabla_a f_h(tu)$ over $t\in[-r,r]$. The lifted credit is $d_ru=uu^\top g+e_ru$, where $|e_r|\le M_3r^2/6$. This expansion also holds when a differentiable surrogate chooses $u$: the leading term is then the canonical reward gradient projected onto the proposed line. More generally, for $u=U_hs$ with isotropic unit $s$, the expected leading term is $U_hU_h^\top g/k$, so one pair measures one direction in the subspace. In RiDEx-Grad, when proposal and evaluation use the same smooth reward and $g\ne0$, setting $u=g/\norm{g}$ yields $d_ru=g+e_ru$; at finite radius, however, the scalar credit can change in magnitude or sign along this line. For a discrete canonical reward, the paired scores still define $d_r$ directly.

For an isotropic random direction $u$, the first-order estimator satisfies
\begin{equation}
 \E[(g^\top u)u]=g/D.
\end{equation}
Consequently, a random-direction control with the same local-loss weight changes both response alignment and expected update magnitude. Dimension rescaling or norm matching defines a different estimator, with corresponding effects on variance and optimizer clipping. Direction quality and estimator normalization thus enter the update through distinct factors: captured response determines the signal, whereas normalization determines its scale before clipping.

\subsection{Orientation invariance and the state-matched update}
Replacing $u$ by $-u$ exchanges the two endpoints, so $d_r(-u)=-d_r(u)$ and
\begin{equation}
 d_r(-u)(-u)=d_r(u)u.
\end{equation}
The symmetric estimator is therefore invariant to the proposal's sign convention. In contrast, shuffling credits across states breaks state correspondence, and reversing credit while fixing the proposal reverses the supervised action.

At a fixed center $h$, the general local loss is $-\widehat c_h^\top U_h^\top b_\theta(h)$, where $b_\theta(h)=(v_\theta(h)-v_h)/q_h$. For RiDEx-Grad, $U_h=[u_h]$ and $\widehat c_h=d_h$. Its parameter gradient is
\begin{equation}
 -\nabla_\theta\Ldir=
 \frac1M\sum_h\frac{1}{q_h}
 \big(\nabla_\theta v_\theta(h)\big)^\top U_h\widehat c_h.
\end{equation}
Let $\Phi_h(w)$ be the terminal reward when the velocity at the fixed center $h$ is replaced by $w$ and the continuation uses frozen behavior parameters. Then $f_h(a)=\Phi_h(v_h+q_ha)$ gives $g=q_h\nabla_w\Phi_h(v_h)$. Write $K_h=D_\theta v_\theta(h)$. In the matched smooth case, the per-center ascent contribution obeys
\[
 \left\|q_h^{-1}K_h^\top(d_ru)-K_h^\top\nabla_w\Phi_h(v_h)\right\|_2
 \le \frac{\norm{K_h}_{\rm op}M_3r^2}{6q_h}.
\]
At $v_\theta(h)=v_h$, the limiting vector recovers the gradient of $\Phi_h(v_\theta(h))$ through the selected velocity prediction. At other replay parameters, the stored credit remains anchored at $v_h$ and is pulled back through the current $K_h$. Differentiating an entire rollout with respect to shared policy parameters also includes contributions from the other velocity predictions; the NFT gradient, joint clipping, and AdamW preconditioning determine the realized step together with the local gradient.

\subsection{Exploration projection and update projection}
The exploration projector $U_hU_h^\top$ acts on the velocity output, whereas RiDEx-CFG's update projector $P_N^\perp=I-G_NG_N^\top/\norm{G_N}^2$ acts on parameter gradients after local responses have been pulled back through their own state Jacobians. Because these are distinct spaces and operations, the rank-two exploration subspace does not constrain all parameter changes to rank two.

For nonzero $G_N$ and $P_N^\perp G_D$, write $e=P_N^\perp G_D/\norm{P_N^\perp G_D}$. The normalized projected combination in the main paper is
\begin{equation}
 G_{\mathrm{CFG}}=\frac{G_N+\eta\norm{G_N}e}{\sqrt{1+\eta^2}}.
\end{equation}
It follows that $\norm{G_{\mathrm{CFG}}}=\norm{G_N}$ and its cosine with $G_N$ is $1/\sqrt{1+\eta^2}$. The trust coefficient therefore controls the added orthogonal component while preserving the combined pre-clipping norm. These identities concern the supplied gradient before optimizer preconditioning; at zero effective local signal the update reduces to $G_N$, and zero norms are handled without division by zero.

Whereas RiDEx-Grad uses $G_N+\lambda G_D$---the gradient of the summed objective with fixed $\lambda$---RiDEx-CFG's gradient-dependent projection and normalization define a combination operator rather than that same fixed-weight loss. The common definition is the basis--credit--combination interface, with the selected operator recorded for each task.

\subsection{Semantic fibers depend on the readout}
For a differentiable multidimensional readout $\Phi$, let $J_h=D(\Phi\circ F_h)(0)$. If the reward locally factors as $f_h(a)=\rho(\Phi(F_h(a)))$, then $g_h=J_h^\top\nabla\rho$, and $\ker J_h\subseteq\ker g_h^\top$. Without that factorization, the reward may change along a readout-invariant direction. Under the regular-rank condition, a readout preimage forms a local level set whose tangent space is the readout Jacobian's kernel. A scalar reward Jacobian, whose rank is at most one by construction, characterizes a different object from a multidimensional semantic readout. Our finite readout sketches analyze this distinction, while RiDEx constructs its exploration direction directly at the generation state.

\subsection{Separating directional and shared endpoint response}
Let $o=(R_+-R_-)/2$ and $e=(R_++R_-)/2-R_0$. Then
\begin{align}
 R_+-R_0&=o+e, & R_--R_0&=-o+e,\\
 \max(R_+,R_-)-R_0&=|o|+e.
\end{align}
The odd term measures orientation-dependent response, while the even term measures the shared displacement from the clean endpoint; their sum gives the positive-branch gain, and $|o|+e$ gives the better branch's gain. Reporting both terms distinguishes relative preference between actions from absolute improvement over clean generation. RiDEx uses the contrast for training, whereas policy evaluation follows a standard trajectory without best-of-two selection.

\section{Implementation and Reproducibility}
\label{supp:implementation}
\subsection{Instantiations and evaluated tasks}
The main method defines RiDEx by an exploration basis $U_h$, a coordinate-credit estimator $\mathcal E_h$, and a combination operator $\mathcal A$. In our experiments, HPSv2 and PickScore use \textbf{RiDEx-Grad}: rank-one reward-gradient proposals, saved-state paired credit, and direct gradient addition (implementation identifier: \texttt{iFANFT-v27}). GenEval, OCR, and the CFG controls use \textbf{RiDEx-CFG}: a two-dimensional condition-anchored basis, paired finite-difference credit, and orthogonal, trust-scaled, norm-matched combination (\texttt{FANFT v5.6r1}). Both preserve the originating state of the local response. The configurations below identify the executed instantiation rather than treating its choices as interchangeable.

\begin{table*}[t]
\centering\small\setlength{\tabcolsep}{5pt}
\begin{tabular}{llll}
\toprule
Task / comparison & Exploration / local update & Endpoints per epoch & NFT updates per epoch\\
\midrule
HPSv2 / PickScore & RiDEx-Grad: rank-one, direct update & 1,152 for both & 1 / 1\\
GenEval learning & RiDEx-CFG: rank-two, projected update & 576 RiDEx / supplementary NFT & 1 / 1\\
OCR learning & RiDEx-CFG: rank-two, projected update & 1152 for both & 1 RiDEx / 4 NFT\\
CFG credit controls & RiDEx-CFG: full / measurement / paired & 576 per arm & 1 per arm\\
\bottomrule
\end{tabular}
\caption{\textbf{Collection and optimization budgets by task.} On HPSv2 and PickScore, DiffusionNFT and RiDEx use the same number of prompts per round and the same number of endpoints per prompt. The supplementary GenEval NFT run is \texttt{ci5cav0s}; primary H200 results use \texttt{iv1oo0cy}. Epochs count collection rounds; the final column specifies the number of NFT updates within each round. Reward fields and proposal construction are detailed below.}
\label{supp:protocols}
\end{table*}

\subsection{Architecture and action collection}
We train Transformer LoRA adapters (Hu et al., 2022) with rank 32 and $\alpha=64$, targeting the attention projections \texttt{add\_k\_proj}, \texttt{add\_q\_proj}, \texttt{add\_v\_proj}, \texttt{to\_add\_out}, \texttt{to\_k}, \texttt{to\_out.0}, \texttt{to\_q}, and \texttt{to\_v}. Text encoders and the VAE remain frozen. Separate current and behavior adapters provide $v_\theta$ and $v_{\bar\theta}$; disabling the adapter gives the pretrained reference velocity.

At each selected center, $q_h=\norm{v_h}_2$ defines the action scale. One velocity prediction is replaced by $v_h+q_h a$, and subsequent predictions use the frozen behavior model. To preserve DPM2's multistep history, each branch starts from the same initial latent with a fresh solver and deterministic prefix, and replay checks verify the saved intervention state and velocity. Computational work includes both the recomputed prefixes and the action-gradient evaluations used to construct RiDEx-Grad proposals. The time-to-target comparison uses aggregate H200 GPU hours (main Fig.~2).

RiDEx-Grad proposals differentiate a continuous scoring path with respect to the injected action, keeping the reward model, VAE, and behavior parameters frozen. HPS uses the HPSv2.1 checkpoint (labeled HPSv2 in figures), while PickScore uses differentiable tensor preprocessing that approximates the canonical processor. The canonical endpoint evaluator independently scores the completed branches and supplies their contrast; a zero-norm proposal produces the zero direction.

\begin{table}[t]
\centering\small
\setlength{\tabcolsep}{4pt}
\begin{tabular}{ll}
\toprule
Setting & RiDEx-Grad value\\
\midrule
Backbone / resolution & SD3.5M / $512\times512$\\
Random-number initialization & 42\\
LoRA rank / $\alpha$ & 32 / 64\\
Collection solver / evaluations & DPM2 / 10\\
Collection CFG & 1\\
Intervention index / radius & 8 (zero-based) / 0.005\\
Local loss weight & 1\\
NFT $\beta$ / advantage clip & 1 / 5\\
Reference-velocity penalty & $10^{-4}$\\
Learning rate / weight decay & $3\times10^{-4}$ / $10^{-4}$\\
AdamW moments / epsilon & $(0.9,0.999)$ / $10^{-8}$\\
Joint gradient clip norm & 1\\
Updates per collection round & 1\\
NFT time views per endpoint & 10\\
Text sequence length & 128\\
Evaluation solver / steps / CFG & Euler / 40 / 4.5\\
Evaluation adapter & current raw adapter\\
Save / evaluation interval & 10 updates\\
Shared comparison update & HPSv2: 210; PickScore: 290\\
\bottomrule
\end{tabular}
\caption{\textbf{RiDEx-Grad configuration for HPSv2 and PickScore.} Shared checkpoints summarize overlapping trajectories; peak scores are selected separately from recorded evaluations. Collection-policy averaging and raw-adapter evaluation serve separate roles.}
\label{supp:config}
\end{table}

\subsection{All-endpoint NFT objective for RiDEx-Grad}
For reward $R_i$ and prompt $c_i$, the round-level advantages and optimality labels are
\begin{equation}
 A_i=\frac{R_i-\operatorname{mean}_{j:c_j=c_i}R_j}
 {\operatorname{std}_{j\in\mathrm{round}}R_j+10^{-4}},
 \quad y_i=\tfrac12+\tfrac1{10}\clip(A_i,-5,5).
\end{equation}
Clean, positive, and negative endpoints all contribute to these statistics. Each endpoint is re-noised at the ten collection-grid times in randomized order, using independent Gaussian draws from the stored seed schedule. With $z_t=(1-t)z_0+t\epsilon$, define
\begin{align}
 v^+_\theta&=\beta v_\theta+(1-\beta)v_{\bar\theta},\\
 v^-_\theta&=(1+\beta)v_{\bar\theta}-\beta v_\theta,
 \qquad \widehat z_0^\pm=z_t-tv^\pm_\theta.
\end{align}
The detached normalization weights are
\begin{equation}
 w_i^\pm=\sg\!\left[\max\!\left(10^{-5},
       \operatorname{mean}|\widehat z_{0,i}^\pm-z_{0,i}|\right)\right],
\end{equation}
evaluated in double precision. Averaging over latent coordinates, the per-sample policy term is
\begin{align}
 \ell_i^{\mathrm{pol}}=\frac5\beta\Bigg[
 &y_i\frac{\operatorname{mean}[(\widehat z_{0,i}^+-z_{0,i})^2]}{w_i^+}\nonumber\\
 &+(1-y_i)\frac{\operatorname{mean}[(\widehat z_{0,i}^--z_{0,i})^2]}{w_i^-}
 \Bigg].
\end{align}
The full NFT loss adds the reference-velocity penalty $10^{-4}\operatorname{mean}[(v_\theta-v_{\mathrm{ref}})^2]$ and averages over endpoints and time views, where the logged field \texttt{kl\_div} denotes the reference-prediction mean-squared error. Table~\ref{supp:config} gives the RiDEx-Grad optimizer settings, and Table~\ref{supp:protocols} summarizes the collection budgets.

\subsection{RiDEx-Grad optimization and behavior averaging}
For RiDEx-Grad, the NFT and local gradients are evaluated at the same pre-update parameters, with the local loss averaging over all selected centers (including zero-credit centers). We sum the parameter gradients, globally reduce them, jointly clip to norm 1, and apply one AdamW step (Loshchilov and Hutter, 2019). When complete prompt groups are unevenly sharded, denominators compensate for rank averaging.

After update $n$, the behavior adapter is updated via
\begin{equation}
 \bar\theta\leftarrow\gamma_n\bar\theta+(1-\gamma_n)\theta,
 \qquad\gamma_n=\min(0.001n,0.5).
\end{equation}
This moving average defines the collection policy, whereas evaluation uses the current raw adapter independently of behavior averaging.

\subsection{Reward units and policy evaluation}
Local credit uses canonical training units: HPS cosine, PickScore logit divided by 26, dense GenEval score, and normalized OCR edit similarity. Displayed HPSv2 and PickScore values multiply these units by 100 and 26, leaving the training objective unchanged. In contrast to NFT's normalized labels, the local loss retains reward units, so reward rescaling changes the relative strength of local and global gradients.

For the saved-state preference runs, policy evaluation generates independent endpoints with 40-step Euler and CFG 4.5, without paired selection. The row seed is $42+10^6+\mathrm{row}$ and remains fixed across checkpoints, with evaluation histories starting at the first evaluated checkpoint. GenEval learning scores use the \texttt{eval\_reward\_geneval} field, while ordinary accuracy and strict accuracy are separate evaluation fields. Run identifiers, score fields, and evaluation schedules accompany the source data.

\subsection{RiDEx-CFG formulation and controls}
The GenEval/OCR configuration constructs a two-dimensional condition-anchored basis at the first velocity evaluation. Its paired finite-difference estimator uses a deterministic, fold-balanced coefficient design, one symmetric pair per base state, base relative radius 0.1, and configured radius multipliers $[1,1,2,2]$. Four paired centers and sixteen clean endpoints give 24 endpoints per prompt. State identity is reconstructed from the initial latent, so the saved intermediate record of RiDEx-Grad is not required for this first-step injection. Local parameter gradients are orthogonalized against the NFT gradient, scaled by the configured trust ratio 0.1 and effective confidence, and combined with final norm matching as in the main update rule, with logged raw, orthogonal, and combined norms distinguishing these stages. The coefficient estimator includes its implementation-specific normalization, converting physical velocity scale to the normalized-action coordinates used in the shared notation. Collection uses ten DPM2 steps, CFG 1, and two training-noise views; the four primary alignment configurations specify 40 evaluation steps and CFG 4.5. The reported curves follow the evaluation outputs stored with those configurations, including the GenEval reward field and corresponding checkpoint identifiers.

For the GenEval and OCR credit ablations, all three arms (\emph{full credit}, \emph{measurement only}, and \emph{paired collection}) have 19 evaluations over epochs 10--190. Full and measurement-only share the configured collection budget and optimizer, while the local-credit switch and trust setting distinguish their updates. Curve-average reward measures learning progression over the shared evaluation window.

The frozen CFG-role experiment compares equal-norm two-dimensional CFG-anchored and random spaces across 144 states (48 prompts with three latent seeds each, totaling 4,752 generated images). Four held-out probe directions per state test predicted branch signs, where signed branch advantage is $\operatorname{sign}(\widehat d)(R_+-R_-)/2$ relative to the pair mean (an abstention contributes zero; ties are retained). Paired differences are first averaged within each prompt across its directions and latent seeds, then bootstrapped over the 48 prompts. PickScore evaluates the same GenEval-domain images, yielding paired composition and preference readouts for each local intervention.

\subsection{Unified collection and learning procedure}
Algorithm~\ref{alg:ridex} summarizes the complete RiDEx training loop, illustrating how the shared $(U,\mathcal E,\mathcal A)$ interface dispatches local directional supervision alongside global DiffusionNFT updates. Depending on whether the target task targets derivative-free text--image alignment or provides a differentiable reward path, the instantiation selects the corresponding subspace proposal, coordinate credit estimator, and gradient-combination rule.
\begin{algorithm}[t]
\caption{RiDEx with an explicit proposal instantiation}
\label{alg:ridex}
\begin{algorithmic}[1]
\Require Behavior $v_{\bar\theta}$; current $v_\theta$; reward $R$; instantiation $(U,\mathcal E,\mathcal A)$
\State Collect independent clean endpoints and select centers.
\For{each center $h$}
  \State Construct $U_h$ from the configured proposal source.
  \State Choose coefficient probes $s_{hj}$ and radii $r_{hj}$.
  \For{each probe $j$}
    \State Generate $F_h(\pm r_{hj}U_hs_{hj})$ with identical prefix and solver history.
    \State Score $R_{hj}^+,R_{hj}^-$ and compute $d_{hj}$.
    \State Retain both endpoints for NFT.
  \EndFor
  \State Set $\widehat c_h=\mathcal E_h(\{s_{hj},d_{hj}\}_j)$.
  \State Reconstruct or load the originating center and evaluate its coordinate loss.
\EndFor
\State Compute $G_N$ from all endpoints and $G_D$ from the local-coordinate objective.
\If{RiDEx-Grad}
  \State $G\gets G_N+\lambda G_D$.
\Else\Comment{RiDEx-CFG}
  \State $G\gets\mathcal A_{\mathrm{CFG}}(G_N,G_D)$: orthogonal projection, trust scaling, norm matching.
\EndIf
\State Jointly clip $G$, take one optimizer step, and update the collection policy according to the selected configuration.
\end{algorithmic}
\end{algorithm}

\section{Reward Response and Exploration Geometry}
\label{supp:frozen}
\subsection{Matched directions and endpoint measurements}
The cross-domain experiment evaluates 256 states per prompt domain and direction, comprising 214 distinct GenEval texts and 256 each for OCR and preference. We average states that share a prompt so that unique prompt means are weighted equally, and compute 95\% percentile intervals from 2,000 bootstrap resamples of the prompt clusters (cf.\ Agarwal et al., 2021). Rotation and single-factor comparisons use 32 states. Across the eight proposal arms---CFG residual, HPS gradient, PickScore gradient, CLIP-alignment gradient, random, wrong-prompt CFG, shuffled-state, and phase-randomized directions---wrong-prompt CFG changes the proposal condition while holding generation and scoring prompts fixed.

The main terminal figure contains 1,024 endpoint pairs, comprising 2,048 branch images: 256 pairs for each target's prescribed proposal. GenEval and OCR use their CFG direction on the corresponding prompt domain, whereas preference targets use their own gradients on preference prompts. The odd/even decomposition distinguishes directional contrast from the mean shift relative to clean generation. Full, parallel, and complement vectors are normalized separately and evaluated through independent continuations at equal action radius.

\subsection{Proposal orientation and finite-action magnitude}
We compare two directional slopes of the same continuous scoring path with the policy frozen: the center autodiff slope $D_{\rm AD}$ and the centered difference $D_{\rm FD}$ at radius $r$. At $r=0.005$, their signs agree at all 32 states for each preference scorer. Across states, however, the median of $|D_{\rm FD}/D_{\rm AD}|$ is 0.202 for HPSv2 (quartiles 0.068, 0.381) and 0.213 for PickScore (0.062, 0.459), with mean relative magnitude errors of 0.735 and 0.726, respectively. Thus the two slopes share the same orientation in these probes, but their relative weight varies by state. During RiDEx-Grad training, terminal credit is obtained from the canonical evaluator on the completed endpoints. The GenEval and OCR derivative checks use a differentiable CLIP proxy, and zero-action replay reproduces the terminal tensor exactly.

\paragraph{Ranking responses at a different radius.}
For each scorer, we match the same 32 states across all four radii in the continuous-path assay, so that each of the $\binom{32}{2}=496$ unordered state pairs contributes a comparison. The two predictors are the absolute autodiff slope at the center and the absolute centered slope at $r=0.005$. For each test radius, the target is which state exhibits the larger absolute reward gap between its own two endpoints (which at fixed radius is identical to the ordering by centered-slope magnitude). The reference-radius endpoints are excluded from the targets, and each comparison reuses the same centers and recorded directions.

\begin{table}[t]
\centering\small\setlength{\tabcolsep}{4.5pt}
\begin{tabular}{lrrrr}
\toprule
Task & $r=.001$ & $.0025$ & $.005$ & $.01$\\
\midrule
GenEval* & 0.108 & 0.185 & 0.271 & 0.376\\
OCR* & 0.137 & 0.210 & 0.276 & 0.351\\
HPSv2 & 0.665 & 0.700 & 0.735 & 0.784\\
PickScore & 0.659 & 0.697 & 0.726 & 0.767\\
\bottomrule
\end{tabular}
\caption{\textbf{Relative magnitude error across the derivative-check radius ladder.} These continuous-path measurements complement the sign-agreement results. Zero-action replay error is zero. *GenEval/OCR use a CLIP proxy. Each analysis has 32 states, with GenEval summaries clustered over 31 distinct prompts.}
\label{tab:numerics}
\end{table}

\begin{table}[t]
\centering\small\setlength{\tabcolsep}{4pt}
\begin{tabular}{lrrr}
\toprule
Task & Sign (\%) & Error $\leq5\%$ (\%) & $c_{\rm norm}$\\
\midrule
GenEval* & 96.8 & 19.4 & 19.31\\
OCR* & 93.8 & 31.2 & 8.93\\
HPSv2 & 100.0 & 0.0 & 161.86\\
PickScore & 100.0 & 0.0 & 61.18\\
\bottomrule
\end{tabular}
\par\smallskip
\begin{tabular}{lrrrr}
\toprule
 & \multicolumn{2}{c}{HPSv2} & \multicolumn{2}{c}{PickScore}\\
\cmidrule(lr){2-3}\cmidrule(lr){4-5}
Test radius & Autodiff & Paired & Autodiff & Paired\\
\midrule
0.0010 & 130 & 471 & 174 & 465\\
0.0025 & 136 & 481 & 175 & 478\\
0.0100 & 139 & 470 & 183 & 484\\
\bottomrule
\end{tabular}
\caption{\textbf{Testing credit magnitude on frozen states.} The upper panel reports sign agreement and 5\% magnitude accuracy at $r=0.005$, with the task/random gradient norm ratio $c_{\rm norm}$ from a separate check. Asterisks mark CLIP proxies. The lower panel counts correctly ordered absolute reward gaps among 496 state pairs per scorer. Paired predictions use $r=0.005$.}
\label{supp:calibration-full}
\end{table}

\begin{figure*}[t]
\centering\includegraphics[width=\textwidth]{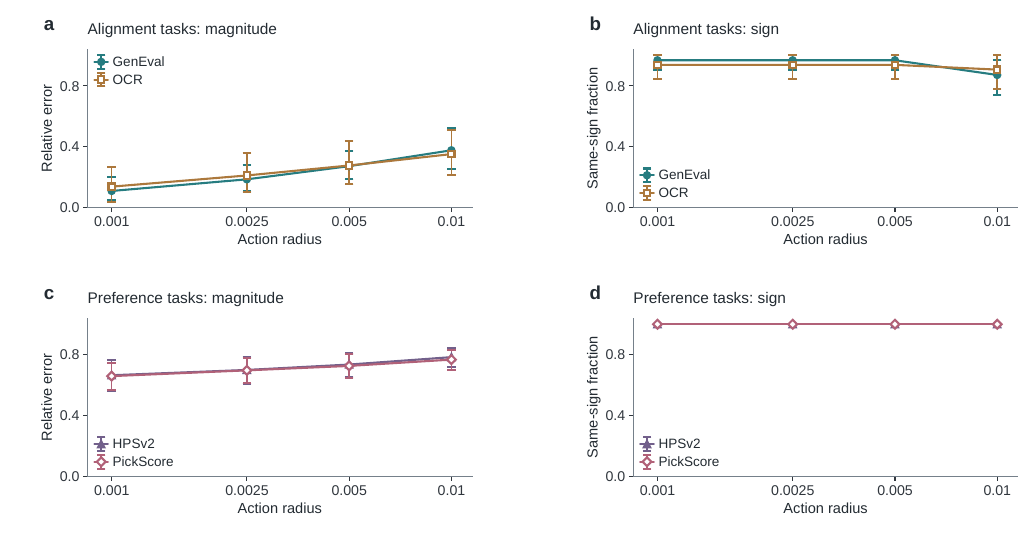}
\caption{\textbf{Directional slopes along the continuous scoring path.} The left panels show relative magnitude error between autodiff and centered differences across four radii. The right panels show sign agreement. Each task has 32 states, with 31 distinct GenEval prompts. Intervals are aggregated by prompt.}
\label{supp:e0}
\end{figure*}

\subsection{Finite readout geometry and state dependence}
CLIP and DINOv2 sketches characterize action-to-readout response at widths 32 and 64. Effective rank describes the response captured at each sketch width, and its growth shows how the sampled representation expands with capacity. Eight of 512 sketches have numerically trimmed ranks; high-mode means aggregate states retaining that mode, and cumulative energy uses the total energy of each finite sketch as its denominator.

\begin{table}[t]
\centering\small\setlength{\tabcolsep}{2.5pt}
\begin{tabular}{llrrr}
\toprule
Task & Basis & Eff. rank & Captured & Transfer\\
 & & $32/64$ & $32/64$ & F/R\\
\midrule
GenEval & CLIP & 31.2/62.2 & 0.28/0.34 & 0.05/0.67\\
GenEval & DINO & 32.0/64.0 & 0.52/0.68 & 0.08/0.81\\
OCR & CLIP & 32.0/64.0 & 0.25/0.28 & 0.05/0.64\\
OCR & DINO & 32.0/64.0 & 0.46/0.63 & 0.06/0.79\\
HPSv2 & CLIP & 31.6/62.7 & 0.24/0.29 & 0.12/0.64\\
HPSv2 & DINO & 32.0/64.0 & 0.44/0.59 & 0.08/0.82\\
PickScore & CLIP & 31.6/62.7 & 0.24/0.29 & 0.12/0.64\\
PickScore & DINO & 32.0/64.0 & 0.44/0.59 & 0.08/0.82\\
\bottomrule
\end{tabular}
\caption{\textbf{Local re-estimation improves held-out readout response.} F/R denotes response cosine for frozen/re-estimated bases. Effective rank and captured fraction are reported at sketch widths 32/64. All task/readout pairs show higher response correspondence with state-specific re-estimation.}
\label{tab:geometry}
\end{table}

\begin{figure*}[t]
\centering\includegraphics[width=\textwidth]{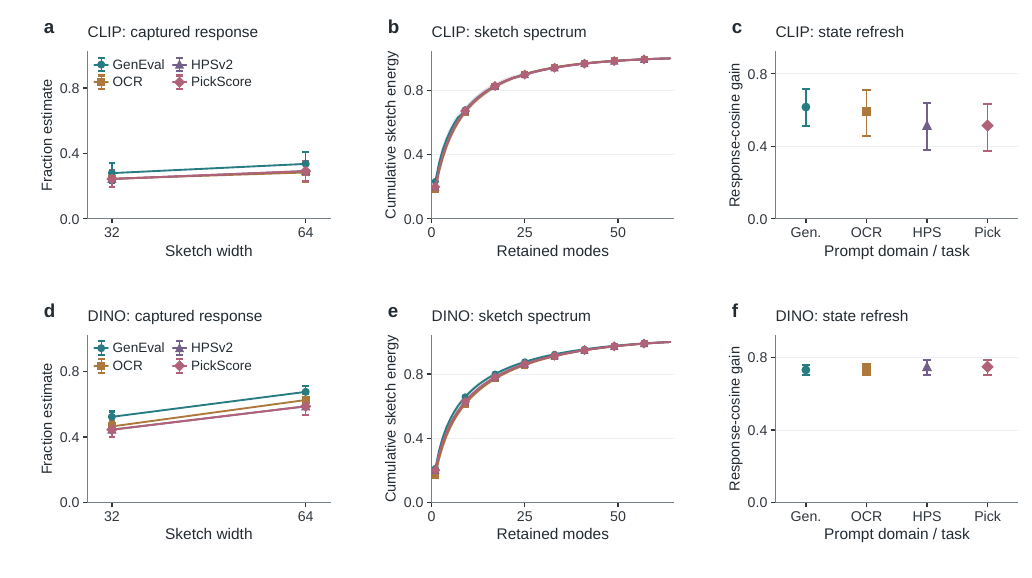}
\caption{\textbf{Finite readout spaces are state-dependent.} Widths 32/64, all retained width-64 spectral positions, and paired frozen-versus-re-estimated basis transfer are shown. Re-estimation raises response cosine by roughly 0.514--0.616 for CLIP and 0.731--0.748 for DINOv2. State-level paired differences are aggregated over 16 prompt clusters per condition. The preference tasks use shared probe states and readouts, so their spectra describe the same local readout geometry.}
\label{supp:readout}
\end{figure*}

\subsection{Reward response beyond the readout-sensitive span}
\begin{figure*}[t]
\centering\includegraphics[width=\textwidth]{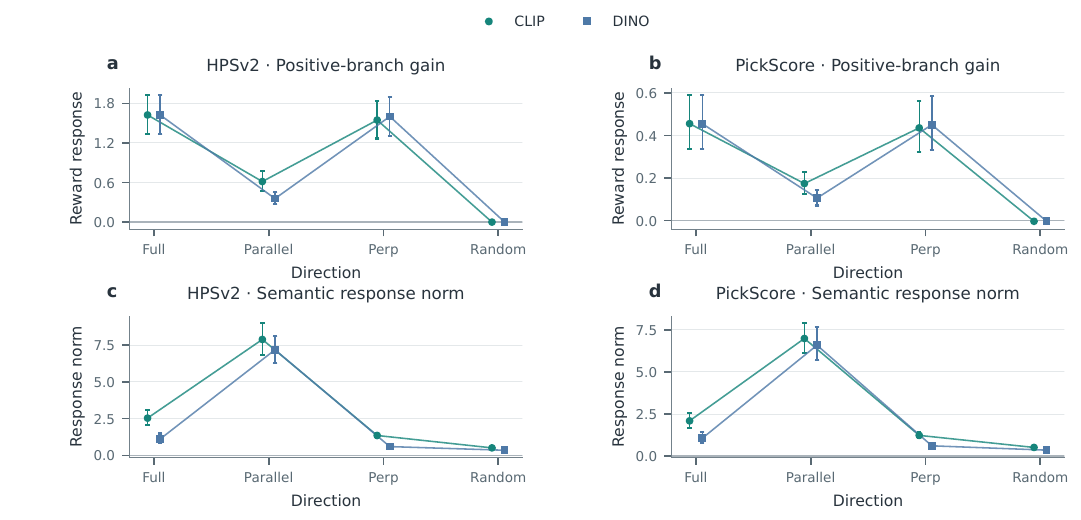}
\caption{\textbf{Readout sensitivity and preference gain favor different directions.} Full, parallel, complement, and random directions have equal radius. Upper panels show reward gains; lower panels show readout-response norms. The two preference tasks and CLIP/DINO bases reveal the response under distinct objectives and readouts.}
\label{supp:e2}
\end{figure*}

For a finite basis $U$, we independently normalize $UU^\top g$ and $(I-UU^\top)g$ to the same action radius. The complement retains substantial preference gain even when the parallel component more strongly changes the embedding. As the orthogonal complement of the measured finite basis, it isolates reward response beyond the captured readout-sensitive span; Figure~\ref{supp:e2} reports both target reward and readout-response norm.

The main sensitivity figure contains all 104 measured readout/task/setting combinations across intervention indices $\{0,2,4,6,8\}$, radii $\{0.001,0.0025,0.01,0.02\}$, and retained ranks $\{1,2,4,8\}$. This sweep isolates three design choices---the state at which to intervene, the displacement magnitude, and the capacity of the proposed readout space---characterizing exploration geometry before parameter optimization.

\section{Policy Learning and Controls}
\label{supp:learning}
\subsection{Reward acquisition and target attainment}
Table~\ref{supp:checkpoint-ledger} summarizes peak rewards and their collection epochs for the supplementary trajectory set.

\begin{table*}[t]
\centering\small\setlength{\tabcolsep}{24pt}
\begin{tabular}{lrr}
\toprule
Task & DiffusionNFT & RiDEx\\
\midrule
GenEval & 0.9432 (70) & 0.9733 (110)\\
OCR & 0.9285 (40) & 0.9683 (30)\\
HPSv2 & 34.2538 (150) & 38.5154 (200)\\
PickScore & 22.6407 (290) & 24.0847 (300)\\
\bottomrule
\end{tabular}
\caption{\textbf{Peak rewards in the supplementary evaluation trajectories.} Entries report each trajectory's maximum reward, with its collection epoch in parentheses. GenEval here uses reference \texttt{ci5cav0s}; the main H200 comparison uses \texttt{iv1oo0cy}. Both retain the evaluation-reward field specified in the implementation details.}
\label{supp:checkpoint-ledger}
\end{table*}

The main H200 comparison and its peak/target tables use GenEval reference run \texttt{iv1oo0cy} through checkpoint 130, paired with RiDEx \texttt{cf4sd7bg} (with measured target 0.950083), whereas the supplementary peak scores in Table~\ref{supp:checkpoint-ledger} use GenEval reference \texttt{ci5cav0s}. Recording both run identifiers distinguishes the primary wall-clock comparison from the supplementary reference trajectory.

\paragraph{H200 time-to-target definition.}
Main Table~2 uses the GPU-hour coordinates of the target-attaining Figure~2 checkpoints. The approximate NFT/RiDEx time pairs are 31.39/12.07 for GenEval, 16.44/8.22 for OCR, 114.63/7.64 for HPSv2, and 222.52/7.67 for PickScore, with speedup ratios computed before rounding. Targets are the baseline's peak measured scores within the respective shared windows, and attainment uses evaluated checkpoints rather than interpolation between them. The time-axis table and extraction metadata accompany the source package.

For shared checkpoints $b_1<\cdots<b_m$ and differences $\Delta_j$, the normalized gap area is
\begin{equation}
 \overline\Delta=\frac1{b_m-b_1}\sum_{j=1}^{m-1}
 \frac{\Delta_j+\Delta_{j+1}}2(b_{j+1}-b_j).
\end{equation}
This mean gap is $+3.8417$ HPSv2 points over epochs 10--210 and $+1.0933$ PickScore points over epochs 10--290, with RiDEx leading DiffusionNFT at all 21/29 exact shared preference checkpoints. For the primary GenEval reference, the analogous gap is $+0.025263$ over epochs 10--130, with positive differences at all 13 shared evaluations. OCR's 20 common evaluations span both rapid acquisition and subsequent retention changes. Each integral uses the shared evaluation horizon, making its averaging interval explicit.

\subsection{Applying and realizing local credit}
Main Table~3 compares \emph{full credit}, \emph{measurement only}, and \emph{paired collection} using the three curves in each panel of main Fig.~5(c--f), where mean AUC is the trapezoidal area divided by the shared epoch span within each task. The GenEval and OCR curves each cover epochs 10--190, while the HPSv2 and PickScore curves cover epochs 10--220 and 10--320, respectively, with one evaluation every ten epochs. The full-minus-measurement differences are $+0.0054145$ for GenEval, $+0.0204805$ for OCR, $+0.7438$ HPSv2 points, and $+0.2297$ PickScore points. On the two alignment tasks, this matched control directly compares the complete projected, trust-scaled update against measurement alone.

\paragraph{Signed credit in preference training.}
In the preference runs, $u_h$ points along the differentiable proposal gradient, so a direct ascent step along this line has nonnegative scalar credit, whereas RiDEx-Grad scores both endpoints with the canonical evaluator. For each training round, let $A=\overline{|d_h|}$ denote the recorded mean absolute credit, and let the mean signed credit $D$ be the difference of the mean positive- and negative-branch gains relative to their shared clean endpoints, divided by $2r$ after conversion to canonical reward units. The mean credit assigned against $u_h$ is $\overline{\max(-d_h,0)}=(A-D)/2$, which exceeds $10^{-5}$ in 47/226 HPSv2 and 53/323 PickScore rounds (matching the counts at thresholds $10^{-7}$ and $10^{-6}$, with equal numbers of centers per round). Summing the negative credit over rounds and dividing by total absolute credit gives just 0.0181\% for HPSv2 and 0.0129\% for PickScore; a positive $(A-D)/2$ identifies at least one reversed label in that round.

\begin{figure*}[t]
\centering\includegraphics[width=\textwidth]{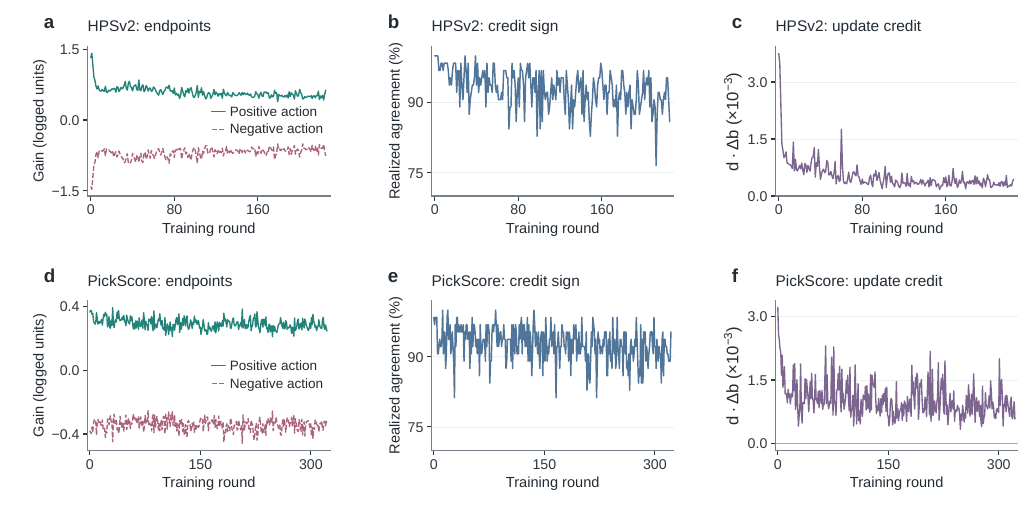}
\caption{\textbf{RiDEx-Grad preference feedback persists throughout the recorded training runs.} All 226 HPSv2 and 323 PickScore events contribute terminal gains, realized credit-sign agreement, and saved-state credit change. The final-30-round sign agreements are 90.73\% and 91.77\%. Credit-sign agreement and credit change measure the post-update action at the saved state. Terminal gains use display units (HPSv2 $\times100$; PickScore $\times26$). Credit and update projections retain canonical training units.}
\label{supp:credit-chain}
\end{figure*}

\begin{figure*}[t]
\centering\includegraphics[width=\textwidth]{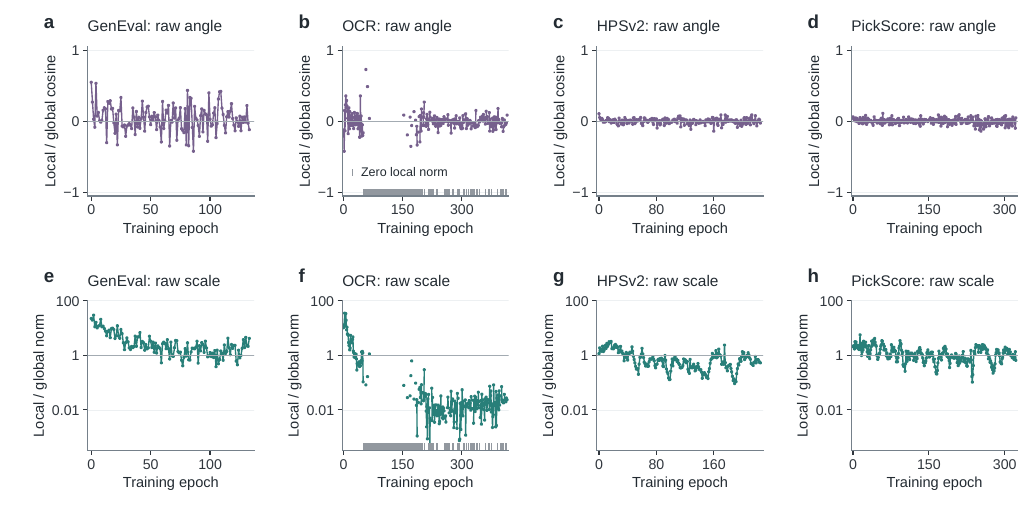}
\caption{\textbf{Local and global gradients contribute distinct directions.} The four tasks contain 134/416/226/323 training events (1,099 total). Top: raw local/global gradient cosine; bottom: raw norm ratio. Gray rugs mark the 177 zero-local-norm OCR events, for which cosine is undefined. CFG-variant statistics precede projection, trust scaling, and norm matching; preference statistics precede gradient addition and Adam. Norm ratios describe these pre-processing gradients; the optimizer applies the task-specific combination rule.}
\label{supp:grad-geometry}
\end{figure*}

Across the first and last 30 preference rounds, the mean stored parameter-step norm changes from 0.470 to 0.157 for HPSv2 and from 0.479 to 0.198 for PickScore, at logged learning rate $3\times10^{-4}$, while the corresponding reference-prediction penalties (\texttt{kl\_div}) change from 0.073 to 0.144 and from 0.051 to 0.116. Together, step norm and reference-prediction penalty characterize the evolving update scale and deviation from the reference velocity.

\subsection{Clean-only and random-pair controls}
In the controlled engine, \texttt{A0} is clean-only NFT (observed through epoch 100 on all tasks) and \texttt{A1} adds random paired endpoints without local credit (observed through epoch 70 on GenEval and 80 on OCR). These controls isolate the effect of adding random paired samples to endpoint learning; Figure~\ref{supp:controls} shows reward progression and optimization dynamics over their evaluated horizons.

\begin{figure*}[p]
\centering
\begin{minipage}{\textwidth}
\captionsetup{type=table,skip=2pt}
\centering\small\setlength{\tabcolsep}{12pt}
\begin{tabular}{lrr}
\toprule
Diagnostic & HPSv2 & PickScore\\
\midrule
Rounds with positive mean credit--action & 226/226 & 323/323\\
Center-wise sign agreement (\%) & 92.98 & 92.71\\
NFT/local gradient cosine & 0.0025 & 0.0017\\
Local gradient norm & 0.0097 & 0.0157\\
NFT gradient norm & 0.0196 & 0.0172\\
\bottomrule
\end{tabular}
\caption{\textbf{State-matched credit is reflected in the realized policy update.} Displacements are measured after the joint optimizer step. Sign agreement is averaged over nonzero-credit centers within each round; gradient statistics are round means. Local and global objectives contribute distinct parameter-update directions.}
\label{tab:update}
\end{minipage}

\par
\captionsetup{skip=2pt}
\includegraphics[width=\textwidth]{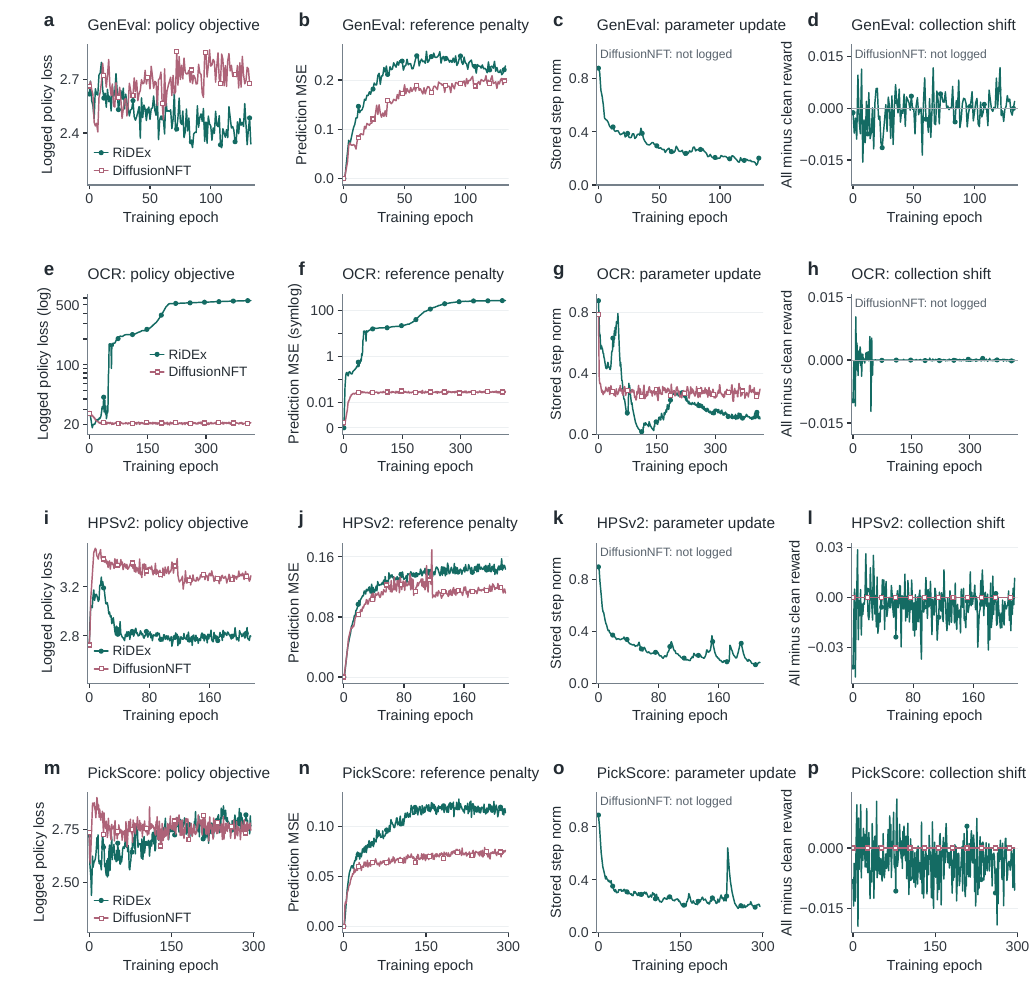}
\caption{\textbf{Optimization scale and collection response across four tasks.} Columns show policy loss, reference-prediction MSE, parameter-step norm, and all-endpoint minus clean-reward gap. The GenEval diagnostic comparator uses the reference run specified in the source data. Shared horizons are 133/415/215/295; marked gaps indicate unavailable fields. OCR NFT updates are averaged within epoch, and log/symlog axes preserve its dynamic range. Objective values are interpreted within their respective loss definitions.}
\label{supp:optimization}
\end{figure*}

\begin{figure*}[t]
\centering\includegraphics[width=\textwidth]{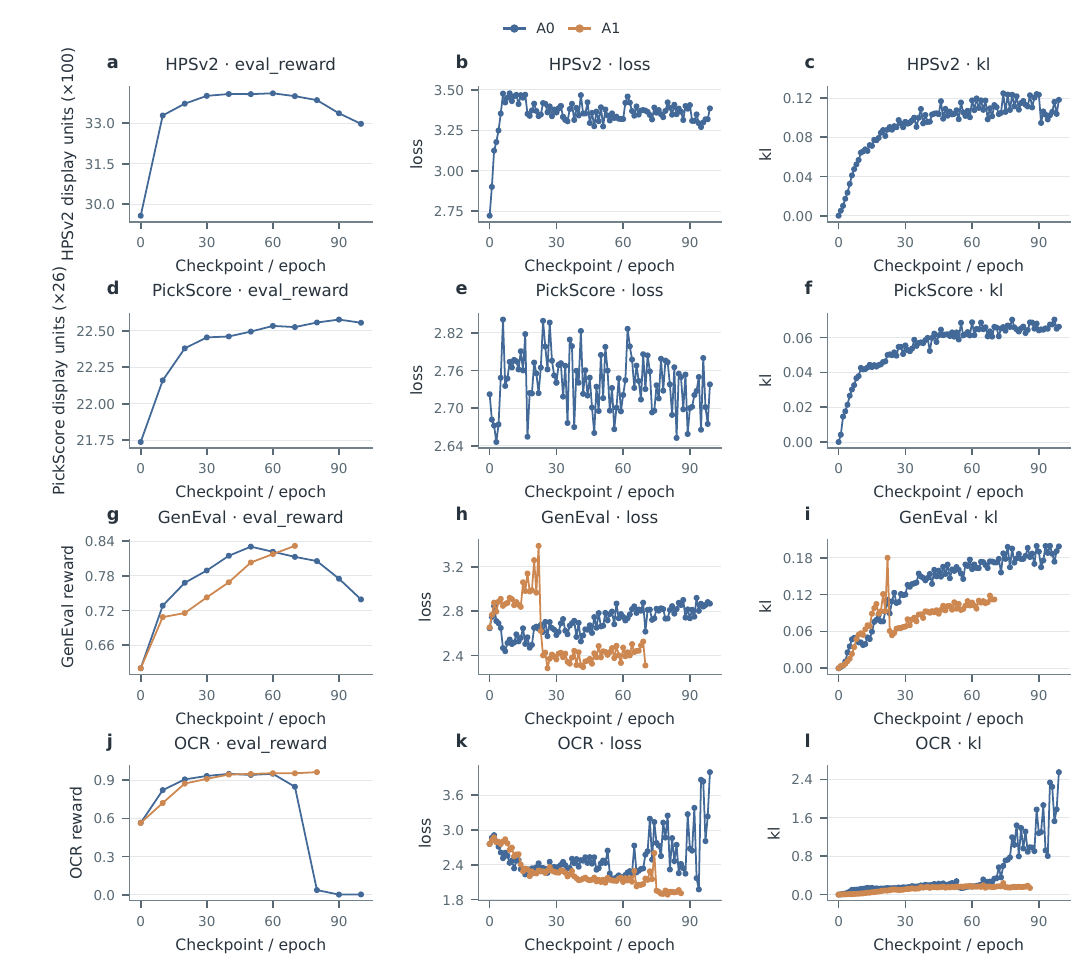}
\caption{\textbf{Random paired collection as an endpoint-learning control.} Rows show reward, policy loss, and reference-prediction penalty. A0 covers all four tasks through epoch 100; A1 covers GenEval through epoch 70 and OCR through epoch 80.}
\label{supp:controls}
\end{figure*}

\clearpage
\onecolumn
\section{Qualitative Comparisons}
\label{supp:qualitative}

\begingroup
\centering
\includegraphics[width=0.90\textwidth]{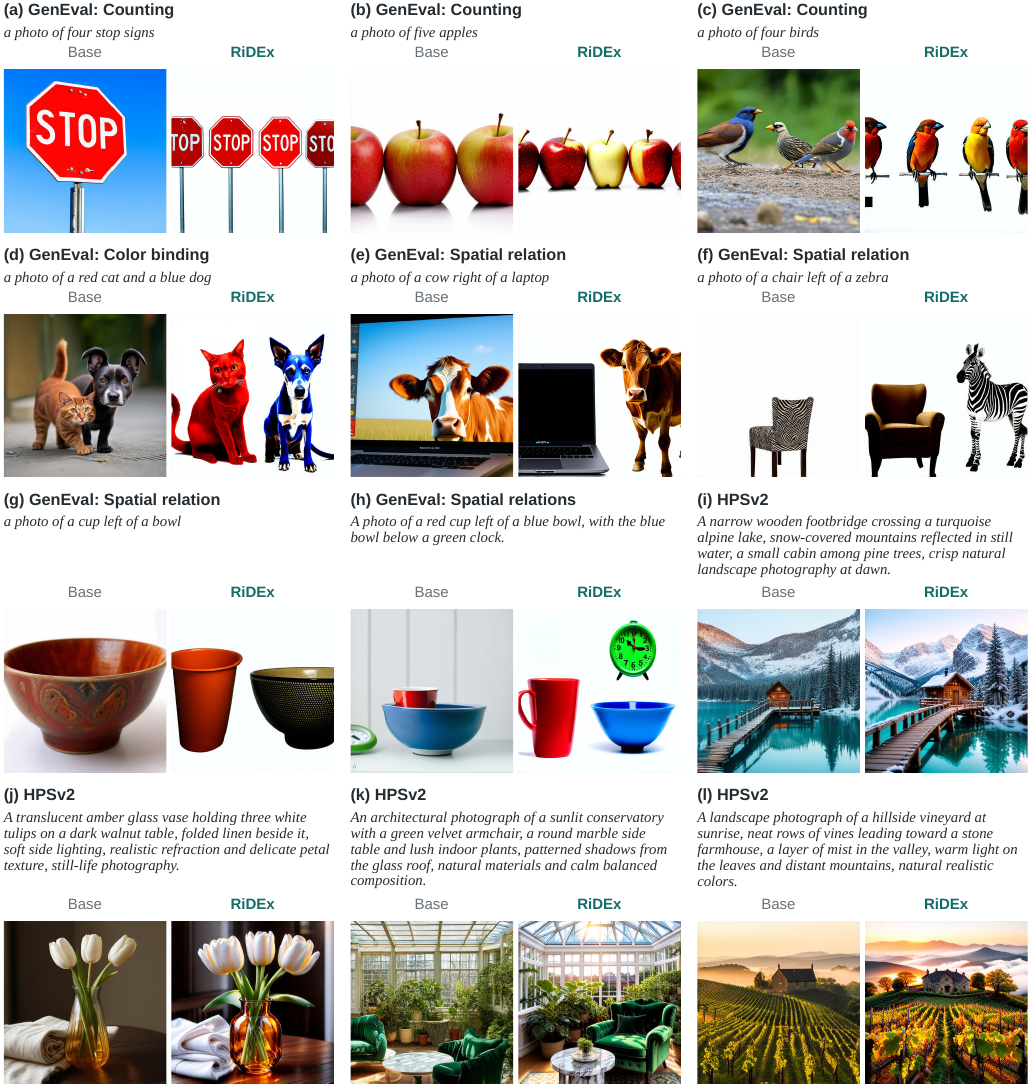}\par
\captionsetup{type=figure,skip=5pt}
\caption{\textbf{Compositional alignment and preference-targeted generation.} (a--h) GenEval examples cover object counting, color binding, and spatial relations; (i--l) HPSv2 examples show landscape, still-life, and architectural scenes. Each pair compares Base (\emph{left}) with RiDEx (\emph{right}) under the prompt printed above.}
\label{supp:qualitative-alignment}
\endgroup
\vspace{6pt}
\noindent\begin{minipage}[t]{0.485\textwidth}
\textbf{Object count and attribute binding.}
In Fig.~\ref{supp:qualitative-alignment}(a,c), RiDEx corrects one stop sign to four and three birds to four, while panel (b) arranges five apples cleanly in a row. Panel (d) binds red to the cat and blue to the dog. In (e,f), the cow appears beside the laptop rather than inside its screen, and the chair and zebra appear as distinct objects rather than a zebra-patterned chair.
\end{minipage}\hfill
\begin{minipage}[t]{0.485\textwidth}
\textbf{Relations and visual appearance.}
Panels (g,h) realize the requested spatial relations: a cup lies left of a bowl, and the red cup, blue bowl, and green clock satisfy both color and position constraints. In (i--l), bridge perspective, amber-glass highlights, sunlit foliage, and vineyard lighting become more vivid, connecting reward-relevant exploration to concrete visual improvements.
\end{minipage}

\newpage
\begingroup
\centering
\includegraphics[width=0.90\textwidth]{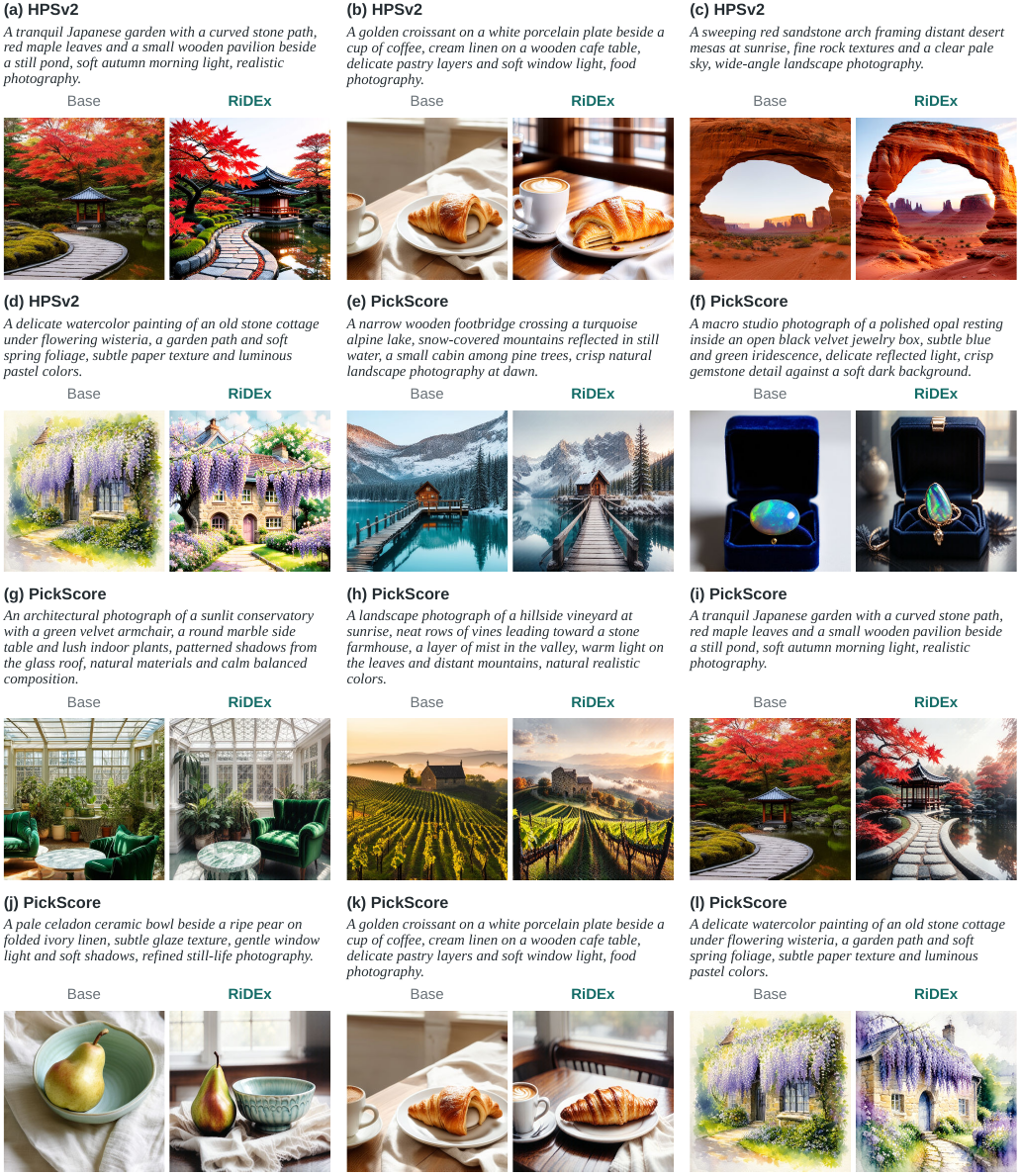}\par
\captionsetup{type=figure,skip=5pt}
\caption{\textbf{Preference-targeted composition, materials, and style.} (a--d) HPSv2 and (e--l) PickScore examples compare Base (\emph{left}) with RiDEx (\emph{right}). Shared garden, food, and cottage prompts across both evaluators illustrate distinct renderings under the two preference objectives.}
\label{supp:qualitative-preference}
\endgroup
\vspace{6pt}
\noindent\begin{minipage}[t]{0.485\textwidth}
\textbf{Composition and material cues.}
In Fig.~\ref{supp:qualitative-preference}(a,c), the garden path and sandstone arch organize the scene around a clear visual route or frame. Panels (b,k) sharpen pastry layers through stronger highlight separation, while (f,j) enhance iridescence and ceramic glaze. In (j), the pear also moves from inside the bowl to beside it, satisfying the stated spatial relation.
\end{minipage}\hfill
\begin{minipage}[t]{0.485\textwidth}
\textbf{Reward-dependent appearance.}
Repeated prompts in (a,i), (b,k), and (d,l) reveal distinct preference-tailored renderings: the HPSv2 cottage emphasizes warm tones and crisp floral detail, whereas PickScore preserves softer watercolor transitions. These visual differences connect task-specific exploration to composition, material, and style, complementing our quantitative analyses.
\end{minipage}

\end{document}